\documentclass[sigconf, screen, numbers]{acmart}

\usepackage{booktabs}
\usepackage{colortbl}
\usepackage{xcolor}
\usepackage{multirow}
\usepackage{hhline}
\usepackage{array}
\usepackage{pifont}
\usepackage{enumitem} %
\usepackage{booktabs}
\usepackage{xcolor}
\usepackage{colortbl}
\usepackage[most]{tcolorbox}
\usepackage{lipsum} %
\usepackage{caption}

\usepackage{hyperref}
\usepackage{url}
\usepackage{wrapfig}
\usepackage{makecell}
\usepackage{longtable} 
\usepackage[figuresleft]{rotating}
\usepackage{listings}
\usepackage[export]{adjustbox}
\usepackage{color}
\usepackage{amsmath}
\usepackage{amsfonts}

\newcolumntype{C}[1]{>{\centering\arraybackslash}p{#1}}
\newcolumntype{L}[1]{>{\arraybackslash}p{#1}}

\usepackage{arydshln}

\usepackage{tcolorbox}

\definecolor{lightgreenzh}{RGB}{0,150,0}
\definecolor{envgreen}{RGB}{50, 140, 80}
\definecolor{lightgrey}{RGB}{247, 247, 247}
\definecolor{lightgray}{RGB}{0.9, 0.9, 0.9}
\definecolor{darkorange}{RGB}{255, 140, 0}
\definecolor{darkblue}{RGB}{84, 112, 198}
\definecolor{lightgreen}{RGB}{145, 204, 117}
\definecolor{lightyellow}{RGB}{250, 200, 88}
\definecolor{lightred}{RGB}{238, 102, 102}
\definecolor{lightblue}{RGB}{115, 192, 222}

\newtcolorbox{promptbox}[2][Prompt]{
  colback=black!5!white,
  arc=5pt,
  boxrule=0.5pt,
  fonttitle=\bfseries,
  title=#1,
  before upper={\small}, fontupper=\fontfamily{ptm}\selectfont,
  colframe=#2, %
}

\definecolor{checkcolor}{RGB}{0,170,0}
\definecolor{xcolor}{RGB}{255,0,0}

\newcommand{\cmark}{\textcolor{checkcolor}{\ding{51}}}
\newcommand{\xmark}{\textcolor{xcolor}{\ding{55}}}

\AtBeginDocument{%
  }

\begin{document}

\title{SVGenius: Benchmarking LLMs in SVG Understanding, Editing and Generation}

\author{Siqi Chen$^{*}$, Xinyu Dong$^{*}$, Haolei Xu, Xingyu Wu, Fei Tang, Hang Zhang, Yuchen Yan, Linjuan Wu, Wenqi Zhang, Guiyang Hou, Yongliang Shen, Weiming Lu, Yueting Zhuang}
\email{{siqichen, syl}@zju.edu.cn}
\affiliation{%
  \institution{Zhejiang University}
  \country{Hangzhou, China}
}
\thanks{$^*$Both authors contributed equally to this research.}

\renewcommand{\shortauthors}{Chen et al.}



\begin{teaserfigure}
  \includegraphics[width=1.0\linewidth]{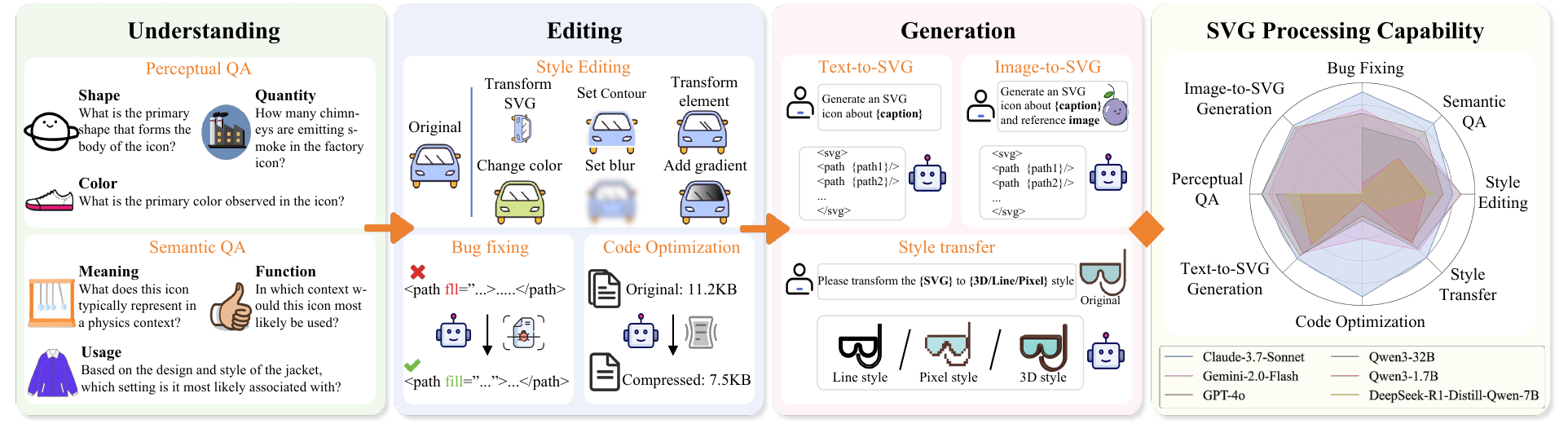}
    \caption{\textbf{Overview of SVGenius}. SVGenius evaluates (M)LLMs capabilities across three progressive dimensions: Understanding (perceptual and semantic QA), Editing (bug fixing, code optimization, style editing), and Generation (text-to-SVG, image-to-SVG, style transfer). Built on real-world data from 24 domains with systematic complexity stratification, our benchmark enables comprehensive assessment of SVG processing capabilities. The radar chart shows representative model performance patterns, revealing distinct capability boundaries.}
    \label{fig:overview}
\end{teaserfigure}

\begin{abstract}
Large Language Models (LLMs) and Multimodal LLMs have shown promising capabilities for SVG processing, yet existing benchmarks suffer from limited real-world coverage, lack of complexity stratification, and fragmented evaluation paradigms. We introduce SVGenius, a comprehensive benchmark comprising 2,377 queries across three progressive dimensions: understanding, editing, and generation. Built on real-world data from 24 application domains with systematic complexity stratification, SVGenius evaluates models through 8 task categories and 18 metrics. We assess 22 mainstream models spanning different scales, architectures, training paradigms, and accessibility levels. Our analysis reveals that while proprietary models significantly outperform open-source counterparts, all models exhibit systematic performance degradation with increasing complexity, indicating fundamental limitations in current approaches; however, reasoning-enhanced training proves more effective than pure scaling for overcoming these limitations, though style transfer remains the most challenging capability across all model types. SVGenius establishes the first systematic evaluation framework for SVG processing, providing crucial insights for developing more capable vector graphics models and advancing automated graphic design applications. Appendix and supplementary materials (including all data and code) are available at \url{https://zju-real.github.io/SVGenius}.

\end{abstract}
\maketitle

\begin{table*}[htbp]
\centering
\small
\caption{\textbf{Comparison of SVGenius with existing SVG processing benchmarks.} We compare construction methods, domain diversity, complexity metrics (paths and control points), and task coverage. SVGenius provides the first comprehensive evaluation across Understanding, Editing, and Generation with systematic complexity stratification. Task abbreviations: PQA (Perceptual QA), SQA (Semantic QA), BF (Bug Fixing), CO (Code Optimization), SE (Style Editing), TTS (Text-to-SVG), ITS (Image-to-SVG), ST (Style Transfer).}
\label{tab:svg_benchmarks}
\begin{tabular}{l c c c c >{\columncolor{gray!10}}c >{\columncolor{gray!10}}c >{\columncolor{gray!10}}c >{\columncolor{gray!10}}c >{\columncolor{gray!10}}c >{\columncolor{gray!10}}c >{\columncolor{gray!10}}c >{\columncolor{gray!10}}c}
\toprule
\multirow{2}{*}{Benchmark} & \multirow{2}{*}{Construct-Method} & \multirow{2}{*}{Domain-Diversity}& \multicolumn{2}{c}{Complex} & \multicolumn{2}{c}{\cellcolor{gray!10}Understanding} & \multicolumn{3}{c}{\cellcolor{gray!10}Editing} & \multicolumn{3}{c}{\cellcolor{gray!10}Generation} \\
\cmidrule(lr){4-5} \cmidrule(lr){6-7} \cmidrule(lr){8-10} \cmidrule(lr){11-13}
&  & &  Paths & Points & PQA & SQA & BF & CO & SE & TTS & ITS & ST \\
\midrule
SVGEditBench~\cite{nishina2024svgeditbench} & Automated & Single & 6.12 & 337.22 & \xmark & \xmark & \xmark & \xmark & \cmark & \xmark & \xmark & \xmark \\
SVGEditBench V2~\cite{nishina2025svgeditbench} & Hybrid & Single & 9.96 & 205.66 & \xmark & \xmark & \xmark & \xmark & \cmark & \xmark & \xmark & \xmark \\
MMSVG-2M~\cite{yang2025omnisvg} & Automated & Multi & - & -  & \xmark & \xmark & \xmark & \xmark & \xmark & \cmark & \cmark & \xmark \\
VGBench~\cite{zou2024vgbench} & Hybrid & Multi & 5.64 & 414.76 & \cmark & \cmark & \xmark & \xmark & \xmark & \cmark & \xmark & \xmark \\
SVGBench~\cite{rodriguez2023starvector} & Automated & Single & 6.08 & 512.54 & \xmark & \xmark & \xmark & \xmark & \xmark & \xmark & \cmark & \xmark \\
Image-Text Bridging~\cite{cai2023leveraging} & Hybrid & Single & - & - & \cmark & \cmark & \xmark & \xmark & \cmark & \xmark & \xmark & \xmark \\
ColorSVG-100K~\cite{chen2025svgbuilder} & Automated & Multi & 13.22 & 952.14 & \xmark & \xmark & \xmark & \xmark & \xmark & \cmark & \cmark & \xmark \\
SVG-Taxonomy~\cite{xu2024exploring} & Automated & Single & 3.98 & 55.58 & \xmark & \xmark & \xmark & \xmark & \cmark & \xmark & \xmark & \xmark \\
SGP-Bench~\cite{qiu2024can} & Hybrid & Multi & 5.50 & 334.38 & \cmark & \cmark & \xmark & \xmark & \xmark & \xmark & \xmark & \xmark \\
\midrule
\textbf{SVGenius(Ours)} & \textbf{Hybrid} & \textbf{Multi} & \textbf{10.15} & \textbf{673.64} & \cmark & \cmark & \cmark & \cmark & \cmark & \cmark & \cmark & \cmark \\
\bottomrule
\end{tabular}
\end{table*}

\section{Introduction}

Scalable Vector Graphics (SVG) has become essential for modern applications ranging from UI design to data visualization, offering lossless scalability and high editability compared to raster formats. However, its XML-based syntax and geometric complexity create significant barriers for non-expert users, motivating automated approaches through differentiable optimization~\cite{jabour831layerwise,jain2023vectorfusion,ma2022towards,xing2024svgdreamer} and deep learning methods~\cite{carlier2020deepsvg,ha2017neural,reddy2021im2vec,tang2024strokenuwa}.

Recent advances in Large Language Models (LLMs)\cite{brown2020language,chowdhery2023palm,touvron2023llama} and Multimodal LLMs\cite{bai2025qwen2,liu2023visual,team2024gemini} have opened new possibilities for SVG processing. These models demonstrate strong capabilities in structured content understanding and generation~\cite{chen2021evaluating,clouatre2019figr,li2022blip}, making them well-suited for SVG manipulation. The textual nature of SVG code naturally aligns with these models' language-centric architecture, enabling direct interpretation and processing of vector graphics. This has led to promising applications across diverse tasks: Iconshop~\cite{wu2023iconshop} and StarVector~\cite{rodriguez2023starvector} focus on icon design and generation, while OmniSVG~\cite{yang2025omnisvg} tackles complex illustration synthesis. However, a critical challenge emerges: \textbf{how do we systematically evaluate and compare the SVG processing capabilities of different models?}

As shown in Table~\ref{tab:svg_benchmarks}, existing benchmarks suffer from three critical limitations: (1) \textbf{limited scope}: reliance on synthetic or overly simplified samples that fail to reflect the structural and semantic diversity of real-world graphics; (2) \textbf{lack of complexity stratification}: uniform treatment of all samples without considering structural complexity that are crucial for understanding model capability boundaries; (3) \textbf{fragmented evaluation}: focus on isolated capabilities rather than comprehensive SVG processing capabilities required for practical applications. Most benchmarks~\cite{nishina2024svgeditbench,nishina2025svgeditbench,yang2025omnisvg,rodriguez2023starvector,cai2023leveraging} evaluate on narrow task subsets, ignoring the multi-stage nature of SVG processing tasks. These gaps hinder fair model comparison and effective guidance for future improvements.

To address these challenges, we introduce SVGenius, a comprehensive benchmark for evaluating SVG processing across multiple dimensions and complexity levels. Built on real-world data from IconFont~\cite{iconfont2024} spanning 24 application domains, our benchmark features a novel complexity stratification framework based on quantitative metrics that organizes samples into three hierarchical levels mirroring real-world design challenges. We evaluate models across three progressive dimensions that capture the full spectrum of SVG processing capabilities: understanding (perceptual and semantic comprehension), editing (code manipulation and optimization), and generation (SVG synthesis and style transfer).

Our benchmark comprises 2,377 queries across 8 task categories and 18 evaluation metrics, enabling systematic assessment of model capabilities. We conduct extensive experiments on 22 mainstream models spanning different scales, architectures, training paradigms, and accessibility (open-source vs. proprietary). Our analysis reveals four key findings: 
(1) substantial performance gaps exist between open-source and proprietary models, with the gap widening on complex tasks; (2) all model families exhibit systematic performance degradation as SVG complexity increases, indicating fundamental limitations in current approaches; (3) reasoning-enhanced training strategies significantly improve performance, particularly for generation tasks requiring multi-step planning; and (4) style transfer emerges as the most challenging capability, with even state-of-the-art models achieving modest performance.

Out contributions can be summaried as: (1) We identify key limitations in existing SVG evaluation approaches and propose a comprehensive solution; (2) We introduce SVGenius, the first large-scale, complexity-stratified benchmark for SVG processing with real-world data; (3) We provide extensive evaluation of 22 models, establishing performance baselines and identifying key factors influencing SVG processing capabilities.

\begin{figure*}[htbp]
    \centering
    \includegraphics[width=1\textwidth]{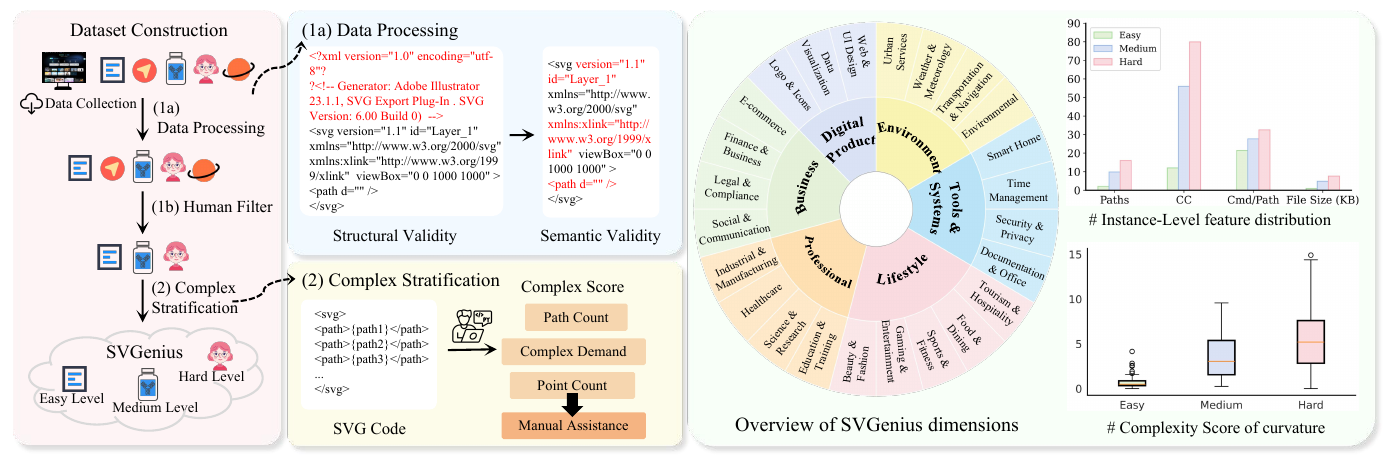}
    \caption{\textbf{SVGenius dataset construction and complexity validation.} Left: systematic pipeline from data collection, processing, human filtering to complexity stratification. Center: 24-domain coverage across diverse applications. Right: validation of complexity stratification showing clear hierarchical separation across Easy, Medium, and Hard levels through feature distributions and complexity scores.}
    \Description{}
    \label{fig:data_construct}
\end{figure*}

\section{Related Works}

\subsection{SVG Processing}
Early AI-driven SVG processing focused on generation using RNNs~\cite{ha2017neural,hu2024supersvg,shen2021clipgen,song2023clipvg,reddy2021im2vec}, VAEs~\cite{lopes2019learned,carlier2020deepsvg,su2023marvel,tang2024strokenuwa,tian2022modern}, and Transformers~\cite{carlier2020deepsvg,wu2023iconshop} to model SVG command sequences. However, these methods faced limitations in representational capacity and geometric consistency for complex graphics. Subsequent work introduced differentiable rasterizers~\cite{li2020differentiable} to bridge vector-raster representation gaps, followed by diffusion-based approaches~\cite{jain2023vectorfusion,thamizharasan2024nivel,xing2023diffsketcher,xing2024svgdreamer,zhang2024text} that improved visual fidelity through iterative refinement.

The emergence of (M)LLMs~\cite{chen2021evaluating,clouatre2019figr,li2022blip,van2008visualizing,vinker2022clipasso,wang2023visionllm} has opened new possibilities for SVG processing. Recent works~\cite{wu2023iconshop,rodriguez2023starvector,yang2025omnisvg,xing2024empowering,wu2024chat2svg} have demonstrated the feasibility of leveraging these models' text understanding capabilities for SVG manipulation, capitalizing on the natural compatibility between XML-based SVG format and language modeling architectures.

\subsection{SVG Benchmarks}
Existing benchmarks evaluate specific aspects of SVG processing capabilities. Image-Text Bridging~\cite{cai2023leveraging} assesses cross-modal perception combining SVG understanding with basic editing. SGP-Bench~\cite{qiu2024can} focuses on semantic consistency in symbolic graphic programs. SVGEditBench~\cite{nishina2024svgeditbench} and V2~\cite{nishina2025svgeditbench} emphasize editing performance using syntactic metrics like MSE and compression ratios. Generation-focused methods include OmniSVG~\cite{yang2025omnisvg}, StarVector~\cite{rodriguez2023starvector}, and SVGBuilder~\cite{chen2025svgbuilder} for text/visual-conditioned synthesis. VGBench~\cite{zou2024vgbench} introduces dual-task evaluation with understanding (VGQA) and generation (VGen) components.

However, existing benchmarks suffer from fragmented evaluation focusing on isolated capabilities, lack of complexity stratification, and reliance on structurally simple samples. SVGenius addresses these limitations by providing a comprehensive framework spanning understanding, editing, and generation with systematic complexity modeling and multi-dimensional evaluation metrics.

\section{Dataset Construction}

Current SVG benchmarks face critical limitations in data quality and diversity. Early efforts like SVGEditBench~\cite{nishina2024svgeditbench} and Image-Text Bridging~\cite{cai2023leveraging} rely on structurally simple icons or synthetic data, while recent works such as OmniSVG~\cite{yang2025omnisvg} and VGBench~\cite{zou2024vgbench} attempt stylistic variation but lack systematic complexity modeling.

To address these limitations, we construct a comprehensive dataset with principled complexity stratification in Figure~\ref{fig:data_construct}. Starting with over 100K SVG samples from Iconfont~\cite{iconfont2024}—a platform of user—created vector icons- we apply systematic processing based on structural validity, semantic effectiveness, and representational compactness. Ten volunteers manually review rasterized versions to ensure semantic clarity. Following standardized preprocessing (geometric normalization, center alignment, attribute standardization), we obtain 927 high-quality SVGs.

\noindent \textbf{Complexity Stratification.} 
We define complexity through three quantitative indicators: path count (structural complexity), control points (geometric intricacy), and complex commands (advanced operations). These metrics are normalized and combined using empirically determined weights. Based on score distributions, we stratify samples into Easy, Medium, and Hard levels (33\%/34\%/33\% split). From each level, we sample 200 candidates, conduct manual visual inspection, and retain 100 high-quality SVGs per level, yielding a balanced 300-sample spanning 24 practical domains dataset with clear quality separation validated in Appendix~\ref{app:data_construct}.

\section{Task Framework}

SVGenius introduces a comprehensive evaluation framework spanning three progressive competency dimensions: Understanding, Editing, and Generation. This design reflects the natural progression of SVG processing skills, from basic comprehension to sophisticated content creation, enabling systematic assessment of model capabilities across the full spectrum of practical applications.

\subsection{Understanding}

Understanding capabilities form the cornerstone of effective SVG processing, encompassing both low-level instruction parsing and high-level semantic construction. Existing benchmarks~\cite{zou2024vgbench,cai2023leveraging,xu2024exploring} oversimplify this dimension through basic question-answer pairs, failing to systematically evaluate multi-level comprehension abilities. We introduce two complementary understanding tasks that progressively assess model capabilities from perceptual recognition to semantic interpretation:

\begin{itemize}[leftmargin=1em,topsep=5pt] 
\item \textbf{Perceptual QA (PQA)} evaluates fundamental visual recognition capabilities essential for SVG interpretation. Models must extract visual cues from SVG code to recognize basic attributes including colors, shapes, spatial relationships, and quantities. Implemented as four-option multiple-choice questions requiring direct code interpretation, this task establishes baseline perceptual processing capabilities using accuracy as the evaluation metric. (cf. Appendix~\ref{app:understanding} for more details)

\item \textbf{Semantic QA (SQA)} advances to higher-level comprehension and abstract reasoning. The task encompasses semantic understanding through three categories: function identification, meaning summarization, and usage prediction. This evaluation requires sophisticated visual-language understanding beyond literal attributes, with accuracy serving as the primary metric for semantic inference capabilities. (cf. Appendix~\ref{app:understanding} for more details)

\end{itemize}
\subsection{Editing}
Building upon understanding foundations, editing tasks assess models' ability to perform precise, structured code manipulations—a critical requirement for practical applications. Current efforts~\cite{nishina2024svgeditbench,nishina2025svgeditbench,cai2023leveraging,qiu2024can} focus narrowly on basic attribute modifications with limited scope diversity.
We design three comprehensive editing scenarios that systematically evaluate 
code manipulation competency:

\begin{itemize}[leftmargin=1em,topsep=5pt] 
\item \textbf{Bug Fixing (BF)} addresses SVG-specific error correction across three primary categories: tag errors (malformed XML structure), attribute errors (incorrect formats), and path command errors (malformed data or sequences). Unlike general program repair benchmarks~\cite{le2015manybugs,lin2017quixbugs,jimenez2023swe,lu2021codexglue}, this task targets unique SVG characteristics requiring both syntactic understanding and semantic preservation. We evaluate the performance of models using repair accuracy to measure the correctness of the fixed output. (cf. Appendix~\ref{app:bf} for more details)

\item \textbf{Code Optimization (CO)} evaluates quality improvement beyond visual correctness. Real-world SVG generation often produces suboptimal code with inefficient structures. The task involves optimizing code following SVGO~\cite{svgo}-inspired principles while preserving rendering output. We evaluate performance using two metrics: mean squared error (MSE) to ensure rendering consistency, and code compression ratio to quantify optimization effectiveness. (cf. Appendix~\ref{app:co} for more details)

\item \textbf{Style Editing (SE)} assesses interactive modification capabilities through six representative operations: global position adjustment, local element movement, set contour, color modification, gradient filling, and blur effects. Since localized modifications yield minimal absolute differences, we introduce relative MSE (rMSE) normalized by reference variance for sensitive quality detection, integrated with existing measures in a four-indicator framework. (cf. Appendix~\ref{app:se} for more details)
\end{itemize}
\subsection{Generation}

Generation represents the most sophisticated capability dimension, requiring models to synthesize complete SVGs from scratch conditioned on natural language or multimodal inputs. Existing benchmarks~\cite{yang2025omnisvg,zou2024vgbench,cai2023leveraging} focus primarily on unimodal scenarios, missing broader challenges of structured construction.
We introduce three progressively challenging generation tasks:

\begin{itemize}[leftmargin=1em,topsep=5pt]

\item \textbf{Text-to-SVG (TTS)} evaluates fundamental natural language to vector graphics translation. To comprehensively assess semantic alignment and code-level differences, we introduce two novel metrics: rCLIP and PSS. These integrate with existing measures to form a three-dimensional framework: (1) Perceptual Quality (HPS~\cite{wu2023human}, Aesthetic~\cite{schuhmann2022aesthetic}) measuring subjective visual appeal, (2) Visual Reproducibility (PSS) assessing code structure consistency, and (3) Semantic Consistency (CLIP~\cite{radford2021learning}, rCLIP) evaluating semantic preservation. (cf. Appendix~\ref{app:tts} for more details)

\item \textbf{Image-to-SVG (ITS)} addresses natural language ambiguity by requiring generation from both images and text. This paradigm mitigates discrepancies between textual descriptions and user expectations by providing visual guidance. Evaluation employs two categories: (1) Perceptual Similarity using LPIPS~\cite{zhang2018unreasonable}, SSIM~\cite{wang2004image}, and DINO~\cite{oquab2023dinov2} for alignment as assessment, and (2) Visual Reproducibility through PSS and MSE for consistency evaluation. (cf. Appendix~\ref{app:mts} for more details)

\item \textbf{Style Transfer (ST)} presents the ultimate challenge, requiring simultaneous content preservation and style adaptation. Despite its importance, standardized SVG style transfer benchmarks remain absent. We introduce a task demanding generation of SVGs retaining structural content while conforming to four predefined stylistic categories. We develop a two-tier automated assessment framework leveraging LLMs to quantify transfer quality from global and local perspectives. (cf. Appendix~\ref{app:st} for more details)

\end{itemize}
\section{Experiment}
\subsection{Experimental Setup}
We evaluate a diverse set of models on SVGenius to assess SVG processing capabilities across different architectures, scales, and training paradigms. Our evaluation encompasses proprietary models (GPT-4o~\cite{hurst2024gpt}, Gemini-2.0-Flash~\cite{team2024gemini}, Claude-3.7-Sonnet~\cite{anthropic2023claude2}), representative open-source models (DeepSeek-R1~\cite{deepseek2024r1}, Qwen2.5/3~\cite{yang2025qwen2,yang2025qwen3}, Llama-3.2~\cite{grattafiori2024llama}, Mistral-Small~\cite{jung2010mistral}), and SVG-specialized systems (Iconshop~\cite{wu2023iconshop}, StarVector~\cite{rodriguez2023starvector}, LLM4SVG~\cite{xing2024empowering}). All models are evaluated under zero-shot settings using default configurations across three complexity levels (Easy, Medium, Hard) with three independent runs per setting for statistical robustness. Due to space constraints, we present results for representative models in the main text, with the complete leaderboard provided in Appendix~\ref{app:full_results}. {In all tables, models are marked as reasoning ({\textcolor{green}{$\star$}}), code ({\textcolor{blue}{$\star$}}), open-source ({\textcolor{red}{$\star$}}), proprietary ({\textcolor{black}{$\star$}}), or specialized ({\textcolor{orange}{$\star$}}) variants.} These metrics (MSE, HPS, PSS, SSIM, LPIPS, DINO) are reported in scientific notation with units of $10^{-2}$ for consistency and readability.

\subsection{Main Results}

We evaluate 22 models across SVGenius spanning understanding, editing, and generation tasks. Tables~\ref{tab:understanding_task}, \ref{tab:edition_task}, \ref{tab:generation} and \ref{tab:image-based_generation_task} present results for representative models, with complete leaderboard in Appendix~\ref{app:full_results}.

\begin{table}[htbp]
\centering
\small
\caption{{Performance on SVG understanding dimension across different model types and difficulty levels for selected models.} Accuracy scores are shown for Perceptual QA and Semantic QA tasks.}
\label{tab:understanding_task}
\begin{tabular}{lC{0.5cm}C{0.5cm}C{0.5cm}C{0.5cm}C{0.5cm}C{0.5cm}}
\toprule
\multirow{2}{*}{Method} & \multicolumn{3}{c}{Perceptual QA(ACC↑)} & \multicolumn{3}{c}{Semantic QA(ACC↑)} \\
\cmidrule(lr){2-4} \cmidrule(lr){5-7}
& \cellcolor{gray!20} \textbf{Easy} & \cellcolor{gray!20} \textbf{Med.} & \cellcolor{gray!20} \textbf{Hard} & \cellcolor{gray!20} \textbf{Easy} & \cellcolor{gray!20} \textbf{Med.} & \cellcolor{gray!20} \textbf{Hard} \\
\midrule
DS-R1-Qwen-32B~\cite{deepseek2024r1}$^{\textcolor{green}{\star}}$ & 64.20 & 43.42 & 42.22 & 51.85 & 52.63 & 37.78 \\
DeepSeek-R1~\cite{deepseek2024r1}$^{\textcolor{green}{\star}}$ & 74.19 & 55.13 & 44.44 & 74.19 & 71.79 & 55.56 \\
Qwen2.5-72B-Ins~\cite{yang2025qwen2}$^{\textcolor{red}{\star}}$ & 67.74 & 38.46 & 24.44 & 50.54 & 47.43 & 51.11 \\
QwQ-32B~\cite{yang2025qwen3}$^{\textcolor{green}{\star}}$ & 62.37 & 34.62 & 33.33 & 53.76 & 57.69 & 37.78 \\
Qwen3-1.7B~\cite{yang2025qwen3}$^{\textcolor{red}{\star}}$ & 46.91 & 22.37 & 28.89 & 41.98 & 44.74 & 22.22 \\
Qwen3-8B~\cite{yang2025qwen3}$^{\textcolor{red}{\star}}$ & 70.96 & 43.59 & 22.22 & 44.09 & 46.15 & 42.22 \\
Qwen3-32B~\cite{yang2025qwen3}$^{\textcolor{red}{\star}}$ & 71.60 & 42.11 & 24.44 & 60.49 & 55.26 & 42.22 \\
Gemini-2.0-Flash~\cite{team2024gemini}$^{\textcolor{black}{\star}}$ & 77.78 & 40.79 & 31.11 & 62.96 & 55.26 & 51.11 \\
GPT-4o~\cite{hurst2024gpt}$^{\textcolor{black}{\star}}$ & 82.72 & 35.53 & 42.22 & 67.90 & 56.58 & 64.44 \\
Claude-3.7-Sonnet~\cite{anthropic2023claude2}$^{\textcolor{black}{\star}}$ & 80.25 & 47.37 & 33.33 & 77.78 & 65.79 & 71.11 \\
\bottomrule
\end{tabular}
\end{table}

\begin{table}[htbp]
\centering
\small
\caption{{Performance on SVG editing dimension across different model types and difficulty levels for selected models.} Results are reported using task-specific metrics (ACC, rMSE, RLD, MSE, CCR) for Bug Fixing, Style Editing, and Code Optimization tasks.}
\label{tab:edition_task}
\begin{tabular}{L{2.7cm}C{0.5cm}C{0.5cm}C{0.5cm}C{0.5cm}C{0.5cm}C{0.5cm}}
\toprule
\multirow{2}{*}{Method} & \multicolumn{1}{c}{Bug F.} & \multicolumn{3}{c}{Style Editing} & \multicolumn{2}{c}{Code Optim.} \\
\cmidrule(lr){2-2} \cmidrule(lr){3-5} \cmidrule(lr){6-7} 
& ACC↑ & ACC↑ & rMSE↑ & RLD↓ & MSE↓ & CCR↑ \\
\midrule
\rowcolor{gray!20}
\multicolumn{7}{c}{\textbf{Easy}} \\
\hline
DS-R1-Qwen-32B~\cite{deepseek2024r1}$^{\textcolor{green}{\star}}$ & 62.63 & 84.81& 75.16 & 1.45 & 5.63 &23.16 \\
DeepSeek-R1~\cite{deepseek2024r1}$^{\textcolor{green}{\star}}$ & 71.00 & 84.62 & 73.66 & 18.21 & 1.01& 23.67 \\
Qwen2.5-72B-Ins~\cite{yang2025qwen2}$^{\textcolor{red}{\star}}$ & 71.00 & 75.95 & 64.25 & 2.08 & 2.98 & 20.57 \\
QwQ-32B~\cite{yang2025qwen3}$^{\textcolor{green}{\star}}$ & 71.43 &91.14 &81.70 &2.61 &3.20&20.20\\
Qwen3-1.7B~\cite{yang2025qwen3}$^{\textcolor{red}{\star}}$ & 22.34 & 74.36 & 67.22 & 33.40 & 11.43 & 35.79 \\
Qwen3-8B~\cite{yang2025qwen3}$^{\textcolor{red}{\star}}$ & 53.00 & 87.34& 80.40&221.88 &2.84 & 18.61\\
Qwen3-32B~\cite{yang2025qwen3}$^{\textcolor{red}{\star}}$ &56.12 &88.46& 82.63& 1.17 &2.55 & 17.96\\
Gemini-2.0-Flash~\cite{team2024gemini}$^{\textcolor{black}{\star}}$ & 69.00 & 86.07& 79.70&  4.78 & 0.72&  20.30 \\
GPT-4o~\cite{hurst2024gpt}$^{\textcolor{black}{\star}}$ & 74.00 & 78.48&65.81 & 46.53 & 1.30 & 10.57 \\
Claude-3.7-Sonnet~\cite{anthropic2023claude2}$^{\textcolor{black}{\star}}$ & 76.00 &79.75&67.17& 20.89 & 0.31& 16.81\\
\midrule
\rowcolor{gray!20}
\multicolumn{7}{c}{\textbf{Medium}} \\
\hline
DS-R1-Qwen-32B~\cite{deepseek2024r1}$^{\textcolor{green}{\star}}$ &46.00 & 56.41 & 53.11 &  140.68 &8.14 &33.10 \\
DeepSeek-R1~\cite{deepseek2024r1}$^{\textcolor{green}{\star}}$ & 63.83 & 62.16 & 53.93 & 1227.70 & 2.02&36.70 \\
Qwen2.5-72B-Ins~\cite{yang2025qwen2}$^{\textcolor{red}{\star}}$ & 50.56 & 59.49& 51.90 & 136.00 &3.41 & 30.49\\
QwQ-32B~\cite{yang2025qwen3}$^{\textcolor{green}{\star}}$ & 46.00& 64.56& 61.27& 41.67 & 5.03&29.28 \\
Qwen3-1.7B~\cite{yang2025qwen3}$^{\textcolor{red}{\star}}$ & 1.22 & 33.90 & 25.55 &  1656.50 &10.20  & 34.87\\
Qwen3-8B~\cite{yang2025qwen3}$^{\textcolor{red}{\star}}$ &35.42 &62.34 &52.10 & 826.21 & 2.63&21.69 \\
Qwen3-32B~\cite{yang2025qwen3}$^{\textcolor{red}{\star}}$ & 46.47 &66.67 &62.54 & 17.13 &2.78 & 26.61\\
Gemini-2.0-Flash~\cite{team2024gemini}$^{\textcolor{black}{\star}}$ & 60.00 &65.82 &60.43 &113.44 & 1.59&27.49 \\
GPT-4o~\cite{hurst2024gpt}$^{\textcolor{black}{\star}}$ & 49.00 &50.63 & 42.64 & 840.73 &4.74 &43.01 \\
Claude-3.7-Sonnet~\cite{anthropic2023claude2}$^{\textcolor{black}{\star}}$ & 75.00 &59.74 &53.61&134.52 &0.86& 26.41 \\

\midrule
\rowcolor{gray!20}
\multicolumn{7}{c}{\textbf{Hard}} \\
\hline
DS-R1-Qwen-32B~\cite{deepseek2024r1}$^{\textcolor{green}{\star}}$ & 34.02 & 55.26& 41.68 & 226.50 & 5.89 &36.41 \\
Deepseek-R1~\cite{deepseek2024r1}$^{\textcolor{green}{\star}}$ & 61.80 & 51.56 & 41.83 & 2610.48 &2.41 &39.95 \\
Qwen2.5-72B-Ins~\cite{yang2025qwen2}$^{\textcolor{red}{\star}}$ & 40.00 & 54.67 & 43.80 & 722.00 & 3.24&36.40 \\
QwQ-32B~\cite{yang2025qwen3}$^{\textcolor{green}{\star}}$ &10.42 &50.00 & 44.04& 533.85&  5.82&12.98 \\
Qwen3-1.7B~\cite{yang2025qwen3}$^{\textcolor{red}{\star}}$ & 0.00& 26.42 &21.94 & 3650.71 & 5.60 & 32.42 \\
Qwen3-8B~\cite{yang2025qwen3}$^{\textcolor{red}{\star}}$ & 25.30& 60.81& 52.21& 935.13 &2.15 &23.68 \\
Qwen3-32B~\cite{yang2025qwen3}$^{\textcolor{red}{\star}}$ &40.86 & 55.26&48.06 & 58.79 &1.50 &26.75 \\
Gemini-2.0-Flash~\cite{team2024gemini}$^{\textcolor{black}{\star}}$ & 53.95& 57.38&42.34 & 628.14 &2.62 &29.35 \\
GPT-4o~\cite{hurst2024gpt}$^{\textcolor{black}{\star}}$ & 51.02 & 42.67& 35.43 & 843.81 & 5.14& 46.91\\
Claude-3.7-Sonnet~\cite{anthropic2023claude2}$^{\textcolor{black}{\star}}$ & 69.00 &52.56&40.75& 27.00 &0.88 &28.27 \\

\bottomrule
\end{tabular}
\end{table}

\noindent \textbf{Proprietary models achieve superior performance but face complexity barriers.} Claude-3.7-Sonnet~\cite{anthropic2023claude2} leads in understanding tasks (80.25\% Easy PQA, 77.78\% Easy SQA) and editing (76\% bug fixing accuracy), while GPT-4o~\cite{hurst2024gpt} excels in generation (20.35 HPS, 19.72 PSS in text-to-SVG, 23.43 PSS in multimodal generation). However, all proprietary models exhibit substantial performance degradation with increasing complexity. GPT-4o~\cite{hurst2024gpt} drops from 82.72\% to 42.22\% in Perceptual QA across difficulty levels, while Gemini-2.0-Flash~\cite{team2024gemini} shows similar degradation from 77.78\% to 31.11\%, indicating fundamental limitations in processing complex SVG structures.

\noindent \textbf{Open-source models reveal significant gaps but show specialized strengths.} Conventional open-source models lag substantially, with Qwen2.5-72B~\cite{yang2025qwen2} achieving 67.74\% Easy PQA compared to Claude-3.7-Sonnet's~\cite{anthropic2023claude2} 80.25\%, representing a 12+\% gap. However, reasoning-enhanced models bridge this gap: DeepSeek-R1~\cite{deepseek2024r1} achieves 74.19\% in both Easy PQA and SQA, closely matching proprietary performance, while QwQ-32B~\cite{yang2025qwen3} excels in editing with 91.14\% style editing accuracy, surpassing most proprietary models including GPT-4o~\cite{hurst2024gpt} (78.48\%). Smaller open-source models face severe limitations, with Qwen3-1.7B~\cite{yang2025qwen3} achieving only 22.34\% bug fixing accuracy and many models completely failing on complex tasks.

\noindent \textbf{Specialized models excel narrowly but lack robustness.} Iconshop~\cite{wu2023iconshop} outperforms most open-source models in text-to-SVG but deteriorates severely on complex tasks (12.95 HPS on Hard) and fails completely at style transfer. StarVector~\cite{rodriguez2023starvector} shows reasonable multimodal performance with competitive SSIM scores (37.60-56.53) but similarly fails other tasks, indicating fundamental limitations in SVG processing. These results show that specialized approaches achieve domain expertise at the cost of general robustness.

\begin{table}[htbp]
\centering
\small
\caption{{Performance on SVG generation dimension across different model types and difficulty levels for selected models.} Results are reported using task-specific metrics (HPS, rCLIP, FSS, Cart., Line, 3D) for Text-based Generation and Style Transfer tasks.}
\label{tab:generation}
\begin{tabular}{{L{2.7cm}C{0.5cm}C{0.5cm}C{0.5cm}C{0.5cm}C{0.5cm}C{0.5cm}}}
\toprule
\multirow{3}{*}{Method} & \multicolumn{3}{c}{Text-to-SVG} & \multicolumn{3}{c}{Style Transfer} \\
\cmidrule(lr){2-4} \cmidrule(lr){5-7}
& HPS↑ & rCLIP↑ & PSS↑ & Cart.↑ & Line↑ & 3D↑ \\
\midrule
\rowcolor{gray!20}
\multicolumn{7}{c}{\textbf{Easy}} \\
\hline
Iconshop~\cite{wu2023iconshop}$^{\textcolor{orange}{\star}}$ & 17.99 & 86.22 & 5.26 & - & - & - \\
LLM4SVG~\cite{xing2024empowering}$^{\textcolor{orange}{\star}}$ & 16.76 & 75.78 & 3.01 & - & - & - \\
DS-R1-Qwen-32B~\cite{deepseek2024r1}$^{\textcolor{green}{\star}}$ & 17.42 & 84.09 & 10.47 & 3.14 & 3.04 & 2.73 \\
DeepSeek-R1~\cite{deepseek2024r1}$^{\textcolor{green}{\star}}$ & 20.37 & 91.39 & 18.07 & 3.13 & 2.68 & 2.87 \\
Qwen2.5-72B-Ins~\cite{yang2025qwen2}$^{\textcolor{red}{\star}}$ & 17.38 & 91.18 & 15.77 & 2.83 & 2.72 & 2.90 \\
QwQ-32B~\cite{yang2025qwen3}$^{\textcolor{green}{\star}}$ & 19.19 & 89.30 & 16.78 & 3.56 & 3.30 & 3.12 \\
Qwen3-1.7B~\cite{yang2025qwen3}$^{\textcolor{red}{\star}}$ & 16.92 & 79.80 & 10.22 & 1.64 & 3.00 & 1.97 \\
Qwen3-8B~\cite{yang2025qwen3}$^{\textcolor{red}{\star}}$ & 18.15 & 85.53 & 12.73 & - & - & - \\
Qwen3-32B~\cite{yang2025qwen3}$^{\textcolor{red}{\star}}$ & 19.39 & 89.13 & 12.62 & 2.93 & 2.92 & 2.80 \\
Gemini-2.0-Flash~\cite{team2024gemini}$^{\textcolor{black}{\star}}$ & 19.82 & 90.96 & 15.94 & 3.17 & 3.16 & 3.25 \\
GPT-4o~\cite{hurst2024gpt}$^{\textcolor{black}{\star}}$ & 20.35 & 90.95 & 19.72 & 3.27 & 2.88 & 2.53 \\
Claude-3.7-Sonnet~\cite{anthropic2023claude2}$^{\textcolor{black}{\star}}$ & 21.35 & 92.90 & 16.69 & 3.68 & 2.08 & 2.93 \\
\midrule
\rowcolor{gray!20}
\multicolumn{7}{c}{\textbf{Medium}} \\
\hline
Iconshop~\cite{wu2023iconshop}$^{\textcolor{orange}{\star}}$ & 14.68 & 77.35 & 3.24 & - & - & - \\
LLM4SVG~\cite{xing2024empowering}$^{\textcolor{orange}{\star}}$ & 14.88 & 68.84 & 2.70 & - & - & - \\
DS-R1-Qwen-32B~\cite{deepseek2024r1}$^{\textcolor{green}{\star}}$ & 15.76 & 75.52 & 7.55 & 1.27 & 2.38 & 3.16 \\
DeepSeek-R1~\cite{deepseek2024r1}$^{\textcolor{green}{\star}}$ & 17.46 & 82.99 & 14.78 & 2.40 & 2.12 & 2.44 \\
Qwen2.5-72B-Ins~\cite{yang2025qwen2}$^{\textcolor{red}{\star}}$ & 16.35 & 79.91 & 13.11 & 3.07 & 2.12 & 3.52 \\
QwQ-32B~\cite{yang2025qwen3}$^{\textcolor{green}{\star}}$ & 16.73 & 77.87 & 17.56 & 3.00 & 2.31 & 3.48 \\
Qwen3-1.7B~\cite{yang2025qwen3}$^{\textcolor{red}{\star}}$ & 16.00 & 70.66 & 11.97 & 1.87 & 1.89 & 1.84 \\
Qwen3-8B~\cite{yang2025qwen3}$^{\textcolor{red}{\star}}$ & 16.70 & 80.37 & 11.26 & - & - & - \\
Qwen3-32B~\cite{yang2025qwen3}$^{\textcolor{red}{\star}}$ & 17.17 & 80.88 & 14.38 & 3.33 & 1.56 & 2.76 \\
Gemini-2.0-Flash~\cite{team2024gemini}$^{\textcolor{black}{\star}}$ & 17.00 & 80.36 & 13.70 & 3.13 & 2.13 & 3.00 \\
GPT-4o~\cite{hurst2024gpt}$^{\textcolor{black}{\star}}$ & 17.51 & 84.72 & 11.97 & 1.67 & 2.00 & 1.56 \\
Claude-3.7-Sonnet~\cite{anthropic2023claude2}$^{\textcolor{black}{\star}}$ & 85.71 & 87.62 & 16.60 & 3.67 & 1.94 & 3.28 \\
\midrule
\rowcolor{gray!20}
\multicolumn{7}{c}{\textbf{Hard}} \\
\hline
Iconshop~\cite{wu2023iconshop}$^{\textcolor{orange}{\star}}$ & 12.95 & 72.55 & 2.12 & - & - & - \\
LLM4SVG~\cite{xing2024empowering}$^{\textcolor{orange}{\star}}$ & 14.62 & 62.98 & 0.02 & - & - & - \\
DS-R1-Qwen-32B~\cite{deepseek2024r1}$^{\textcolor{green}{\star}}$ & 14.56 & 72.71 & 6.56 & 2.20 & 2.08 & 2.40 \\
DeepSeek-R1~\cite{deepseek2024r1}$^{\textcolor{green}{\star}}$ & 16.86 & 83.06 & 10.07 & 2.16 & 2.00 & 2.13 \\
Qwen2.5-72B-Ins~\cite{yang2025qwen2}$^{\textcolor{red}{\star}}$ & 15.51 & 75.22 & 9.57 & 2.64 & 1.96 & 3.20 \\
QwQ-32B~\cite{yang2025qwen3}$^{\textcolor{green}{\star}}$ & 16.36 & 82.24 & 8.45 & 1.92 & 1.92 & 2.40 \\
Qwen3-1.7B~\cite{yang2025qwen3}$^{\textcolor{red}{\star}}$ & 14.52 & 68.91 & 4.46 & 1.32 & 2.24 & 1.20 \\
Qwen3-8B~\cite{yang2025qwen3}$^{\textcolor{red}{\star}}$ & 15.34 & 77.47 & 8.64 & - & - & - \\
Qwen3-32B~\cite{yang2025qwen3}$^{\textcolor{red}{\star}}$ & 16.02 & 81.84 & 9.88 & 2.96 & 1.60 & 3.33 \\
Gemini-2.0-Flash~\cite{team2024gemini}$^{\textcolor{black}{\star}}$ & 16.01 & 78.61 & 9.89 & 1.32 & 1.08 & 1.26 \\
GPT-4o~\cite{hurst2024gpt}$^{\textcolor{black}{\star}}$ & 16.69 & 82.81 & 10.39 & 1.84 & 1.96 & 2.73 \\
Claude-3.7-Sonnet~\cite{anthropic2023claude2}$^{\textcolor{black}{\star}}$ & 18.74 & 87.97 & 10.19 & 2.64 & 1.80 & 3.33 \\
\bottomrule
\end{tabular}
\end{table}

\begin{table}[htbp]
\centering
\small
\caption{Performance on SVG generation dimension across different model types and difficulty levels. Results are reported using task-specific metrics (SSIM, LPIPS, MSE, DINO, PSS) for Image-to-SVG.}
\label{tab:image-based_generation_task}
\begin{tabular}{{L{2.9cm}C{0.6cm}C{0.6cm}C{0.6cm}C{0.6cm}C{0.6cm}}}
\toprule
\multirow{2}{*}{Method} & \multicolumn{5}{c}{Image-to-SVG} \\
\cmidrule(lr){2-6}  
& SSIM↑ & LPIPS↓ & MSE↓ & DINO↑ & PSS↑\\
\midrule
\rowcolor{gray!20}
\multicolumn{6}{c}{\textbf{Easy}} \\
\hline
StarVector(8B)~\cite{rodriguez2023starvector}$^{\textcolor{orange}{\star}}$ & 37.60 & 36.80 & 43.71 & 73.98 & -\\
InternVL3-8B~\cite{zhu2025internvl3}$^{\textcolor{red}{\star}}$ & 46.16 & 44.10 & 27.26 & 76.35 & 2.72 \\
Qwen2.5-VL-3B-Ins~\cite{bai2025qwen2}$^{\textcolor{red}{\star}}$ & 50.66 & 38.91 & 28.32 & 73.92 & 8.15 \\
Qwen2.5-VL-72B-Ins~\cite{bai2025qwen2}$^{\textcolor{red}{\star}}$ & 46.75 & 40.85 & 24.52 & 79.83 & 17.60 \\
Gemini-2.0-Flash~\cite{team2024gemini}$^{\textcolor{black}{\star}}$ &50.07 &37.01 & 21.10& 85.21&19.80 \\
GPT-4o~\cite{hurst2024gpt}$^{\textcolor{black}{\star}}$ & 52.41&33.81 &19.21&87.48 &23.43 \\
Claude-3.7-Sonnet~\cite{anthropic2023claude2}$^{\textcolor{black}{\star}}$ & 54.02&32.12 &17.13 &89.91 & 23.70\\
\midrule
\rowcolor{gray!20}
\multicolumn{6}{c}{\textbf{Medium}} \\
\hline
StarVector(8B)~\cite{rodriguez2023starvector}$^{\textcolor{orange}{\star}}$ & 52.95 & 41.20 & 21.63 & 67.28 & - \\
InternVL3-8B~\cite{zhu2025internvl3}$^{\textcolor{red}{\star}}$ & 45.26 & 48.47 & 19.51 & 73.90 & 1.72\\
Qwen2.5-VL-3B-Ins~\cite{bai2025qwen2}$^{\textcolor{red}{\star}}$ &47.50& 44.60& 22.85& 69.33&1.93 \\
Qwen2.5-VL-72B-Ins~\cite{bai2025qwen2}$^{\textcolor{red}{\star}}$ & 45.29 & 47.93 & 20.04 & 76.41& 15.82 \\
Gemini-2.0-Flash~\cite{team2024gemini}$^{\textcolor{black}{\star}}$ & 46.98& 45.48& 15.65& 79.74& 10.16\\
GPT-4o~\cite{hurst2024gpt}$^{\textcolor{black}{\star}}$ &49.41 & 42.29& 15.44& 81.29& 12.64\\
Claude-3.7-Sonnet~\cite{anthropic2023claude2}$^{\textcolor{black}{\star}}$ &51.36 &37.67 & 12.33& 86.17& 16.58\\
\midrule
\rowcolor{gray!20}
\multicolumn{6}{c}{\textbf{Hard}} \\
\hline
StarVector(8B)~\cite{rodriguez2023starvector}$^{\textcolor{orange}{\star}}$ & 56.53 & 42.12 & 16.47 & 65.81 &- \\
InternVL3-8B~\cite{zhu2025internvl3}$^{\textcolor{red}{\star}}$ & 46.83 & 51.39 & 18.48 & 75.05 &1.77 \\
Qwen2.5-VL-3B-Ins~\cite{bai2025qwen2}$^{\textcolor{red}{\star}}$ & 52.84 & 47.35 & 19.41 & 70.88 & 1.52\\
Qwen2.5-VL-72B-Ins~\cite{bai2025qwen2}$^{\textcolor{red}{\star}}$ & 42.74& 49.86& 17.72& 75.88&11.60\\
Gemini-2.0-Flash~\cite{team2024gemini}$^{\textcolor{black}{\star}}$ &48.29 &46.38 & 12.54& 80.21& 7.30 \\
GPT-4o~\cite{hurst2024gpt}$^{\textcolor{black}{\star}}$ &50.66 &43.40 & 12.77& 81.36&11.67 \\
Claude-3.7-Sonnet~\cite{anthropic2023claude2}$^{\textcolor{black}{\star}}$ &51.15 &39.59 & 11.23& 85.51& 15.47\\
\bottomrule
\end{tabular}
\end{table}

\subsection{Analysis}

\noindent \textbf{Model Scale and Architecture Effects}
Scaling yields substantial improvements within model families: Qwen3 variants improve from 22.34\% to 56.12\% easy bug fixing accuracy when scaling from 1.7B to 32B parameters. 
Multimodal architectures consistently outperform text-only variants in generation, with GPT-4o~\cite{hurst2024gpt} achieving 23.43 vs 19.72 PSS scores, indicating that visual modalities enhance spatial reasoning capabilities essential for SVG generation.

\noindent \textbf{Reasoning-Enhanced Training Improves SVG Processing}
Reasoning augmentation enables models to transcend scale limitations through systematic problem-solving approaches. DS-R1-Qwen-32B~\cite{deepseek2024r1} achieves 51.85\% in Easy SQA despite having fewer parameters than Qwen2.5-72B~\cite{bai2025qwen2} (50.54\%). QwQ-32B similarly outperforms conventional models of similar scale, achieving 91.14\% vs 88.46\% easy style editing accuracy compared to Qwen3-32B~\cite{yang2025qwen3}. These results suggest reasoning training develops better systematic approaches to SVG's hierarchical structure and semantic relationships.

\noindent \textbf{Complexity-Performance Degradation Patterns}
Performance degradation with increasing complexity is universal but varies by task type. Understanding tasks show steep degradation (Claude-3.7-Sonnet~\cite{anthropic2023claude2}: 80.25\% to 33.33\% PQA, GPT-4o~\cite{hurst2024gpt}: 82.72\% to 42.22\%), while editing exhibits moderate drops (10-30\% across difficulty levels for most models). Generation tasks prove most resilient with 5-10\% PSS score declines. Critically, degradation patterns remain consistent across model families—all Qwen variants show similar proportional decreases, and reasoning-enhanced models follow identical trends despite higher baselines, indicating fundamental limitations in current approaches rather than model-specific weaknesses.

\noindent \textbf{Task-Specific Capability Boundaries}
Distinct capability boundaries reveal fundamental SVG processing challenges. Understanding consistently exceeds generation performance: top models achieve 70-80\% semantic understanding but struggle with structural synthesis (PSS scores rarely exceed 20), indicating a comprehension-creation gap. Multimodal inputs enhance PSS scores by 5-10\%, though benefits decrease on complex tasks, suggesting visual guidance supports basic structural comprehension more than intricate geometric synthesis. Style transfer emerges as the most challenging task, with only reasoning-enhanced and proprietary models achieving meaningful adaptation (scores >3.0) while others perform superficial modifications (<2.5).

\noindent \textbf{Failure Mode Analysis}
Analysis of failure patterns reveals scale-dependent weaknesses that explain performance boundaries. Small models (<7B) exhibit fundamental syntactic failures, with Qwen3-1.7B~\cite{yang2025qwen3} achieving only 22.34\% bug fixing accuracy and 0\% on hard tasks. Medium models (7-30B) demonstrate semantic limitations, excelling at local edits but struggling with global manipulations as evidenced by Qwen3-8B's~\cite{yang2025qwen3} inconsistent performance (Style Editing: 87.34\% vs Bug Fixing: 53.00\%). Large models show improved global structural understanding but suffer from style abstraction failures, with even top performers like Claude~\cite{anthropic2023claude2} achieving only 2-4 range scores in style transfer. This progression from syntactic to semantic to abstraction failures indicates that SVG processing requires hierarchical skill development that current training approaches only partially address.

\section{Conclusion}

We introduce SVGenius, the first comprehensive benchmark for evaluating LLMs capabilities in SVG processing, comprising 2,377 queries across understanding, editing, and generation with systematic complexity stratification. Evaluation of 22 models reveals that while proprietary models outperform open-source counterparts, all models degrade with increasing complexity, and reasoning-enhanced training proves more effective than pure scaling. These findings indicate fundamental limitations in current approaches and suggest specialized training methods that better capture vector graphics' structured nature. SVGenius establishes a foundation for advancing SVG processing research and automated graphic design.

\clearpage

\bibliographystyle{ACM-Reference-Format}
\bibliography{main}

\clearpage

\appendix

\section{Overview of SVGenius}
\begin{figure}[h]
    \centering
    \includegraphics[width=0.85\linewidth]{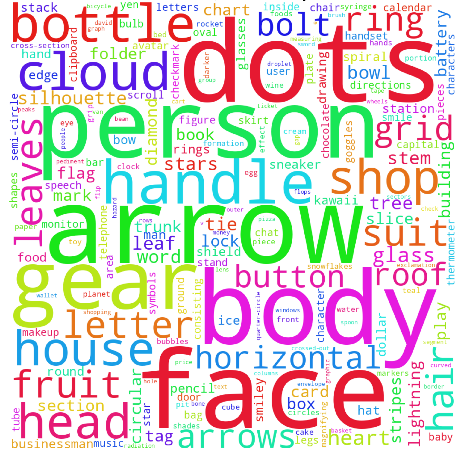}
    \caption{Word cloud visualization of the 24 diverse application domains covered in SVGenius dataset.}
    \Description{}
    \label{fig:wordcloud}
\end{figure}

SVGenius incorporates samples from IconFont~\cite{iconfont2024}, a real-world vector icon repository, to ensure comprehensive coverage of SVG usage patterns and semantic contexts across diverse domains. Our dataset spans six major thematic clusters including Digital Products (encompassing UI/UX design, e-commerce, and web applications), Professional services (including business, finance, legal, and healthcare sectors), Lifestyle categories (covering food, sports, entertainment, and personal items), Work \& Systems (featuring tools, time management, and security applications), Environmental domains (incorporating nature, science, and sustainability themes), and Smart Home technologies (including IoT devices and home automation systems). This comprehensive domain architecture is visualized through a word cloud representation of vector icon captions in Figure~\ref{fig:wordcloud}.

\section{Dataset Quality Assessment}
\label{app:data_construct}
\subsection{Instance-level Complexity Metrics}
Our instance-level complexity quantification employs automated SVG parsing to extract fundamental structural metrics. We compute path count through XML parsing, numerical parameter count via regular expression matching, and complex command frequency by identifying curve-type operations (C, Q, A, S, T variants). Additional metrics include total command count, average commands per path, and file size in bytes to capture geometric density and syntactic complexity. Dataset complexity scores are presented in Table \ref{tab:instance_metrics}.

\begin{table}[htbp]
\centering
\tiny
\caption{Instance-level complexity metrics}
\label{tab:instance_metrics}
\begin{tabular}{lcccccc}
\toprule
Level & Paths & Commands & Points & Complex Commands& Cmd/Path &File Size (KB) \\
\midrule
Easy & 2.14 & 35.84 & 132.7 & 12.03 & 21.39 & 0.96 \\
Medium & 9.87 & 150.93 & 739.56 & 56.01 & 27.68 & 4.83 \\
Hard & 16.02 & 228.87 & 1148.67 & 79.94 & 32.57 & 7.63 \\
\bottomrule
\end{tabular}
\end{table}
\subsection{Structure-Aware Complexity Modeling}
To systematically stratify SVG samples across hierarchical complexity levels, we implement a multi-dimensional scoring framework based on weighted command analysis and advanced structural metrics. Each SVG path command receives complexity weights reflecting computational demands: basic operations (M, L, Z) weighted at 1, quadratic curves (Q, T) at 2-3, cubic curves (C, S) at 3-4, and elliptical arcs (A) at 5. We further compute command type entropy, coordinate precision entropy, and curvature complexity. Final stratification integrates these orthogonal measures through principal component analysis, enabling robust classification into three hierarchical levels that demonstrate statistically significant differences (ANOVA F-tests, p < 0.001) across multiple complexity dimensions.Dataset complexity scores are presented in Table~\ref{tab:complexity_metrics}.
\begin{table}[htbp]
\centering
\small
\caption{Complexity metrics across hierarchical levels}
\label{tab:complexity_metrics}
\begin{tabular}{lcccc}
\toprule
Level & Curvature & Structure & Type Entropy \\
\midrule
Easy & 0.70 & 3.58 & 1.48  \\
Medium & 3.56 & 15.09 & 1.67  \\
Hard & 5.46 & 22.89 & 1.68\\
\bottomrule
\end{tabular}
\end{table}
\subsection{Manual Assessment}
\begin{figure}[htbp]
\centering
\includegraphics[
    width=1\linewidth,
    trim={0pt 0pt 0pt 0pt},
    clip
]{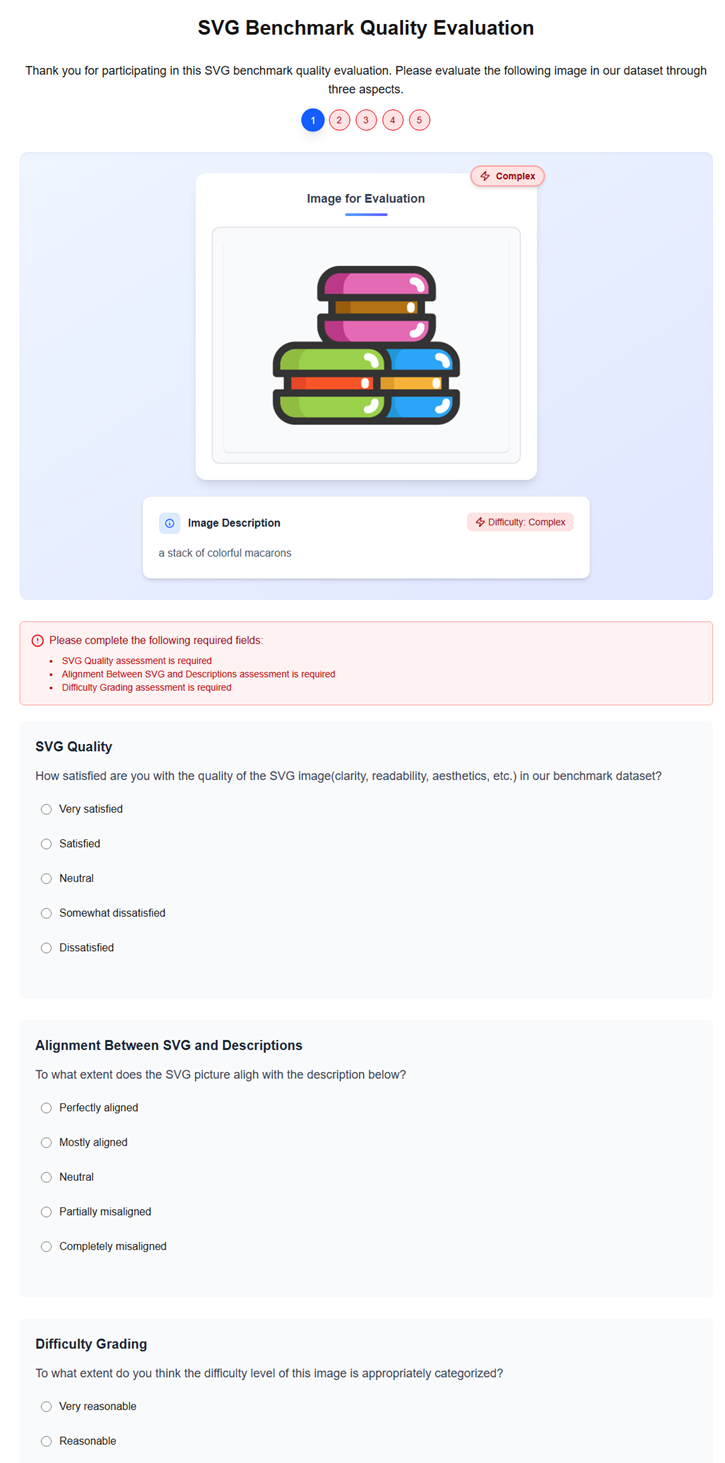}
\caption{Screenshot of Our Questionnaires.}
\label{fig:question}
\vspace{-1.5em}
\end{figure}

To systematically evaluate the quality of our dataset, we conducted a user study involving 10 volunteer participants. As shown in Figure~\ref{fig:question}, the assessment focused on the following three key dimensions:

\begin{itemize} [leftmargin=1em,topsep=5pt]
\item\textbf{Visual Quality}: The aesthetic appeal and clarity of the SVG graphics, including factors such as compositional balance, rendering fidelity, and visual coherence.

\item\textbf{Semantic Alignment}: The consistency between each SVG graphic and its associated image description, evaluating whether the semantic content of the text accurately reflects the visual content.

\item\textbf{Difficulty Stratification}: The appropriateness and discriminative power of the assigned difficulty levels, assessing whether the difficulty annotations reflect meaningful variation in task complexity.

\end{itemize}

This study was designed to provide a holistic evaluation of the dataset's validity across visual presentation, semantic alignment, and difficulty calibration, ensuring that it meets the standards required for reliable model evaluation.

\section{Full Leaderboard}
\label{app:full_results}
For completeness and reproducibility, we provide the full versions of the tables referenced in the main text. Due to space constraints, only abridged versions of these tables were included in the body of the paper. Table~\ref{tab:understanding_task_full}, 
Table~\ref{tab:edition_task_full}, 
 Table~\ref{tab:generation_full} and Table~\ref{tab:style_transfer_task} presented more detailed statistics and results, which may be of interest to researchers seeking to replicate or analyze our findings further. 

\begin{table*}[htbp]
\centering
\small
\caption{Performance on SVG understanding dimension across different model types and difficulty levels. Accuracy scores are shown for Perceptual QA and Semantic QA tasks, with models marked as reasoning ({\textcolor{green}{$\star$}}), code ({\textcolor{blue}{$\star$}}), open-source ({\textcolor{red}{$\star$}}), or proprietary ({\textcolor{black}{$\star$}}) variants.}
\label{tab:understanding_task_full}
\begin{tabular}{l c c c c c c}
\toprule
\multirow{2}{*}{Method} & \multicolumn{3}{c}{Perceptual QA(ACC↑)(\%)} & \multicolumn{3}{c}{Semantic QA(ACC↑)(\%)} \\
\cmidrule(lr){2-4} \cmidrule(lr){5-7}
& Easy & Medium & Hard & Easy & Medium & Hard \\
\midrule
DeepSeek-R1-Distill-Qwen-1.5B~\cite{deepseek2024r1}$^{\textcolor{green}{\star}}$ & 43.01 & 26.92 & 17.78 & 31.18 & 20.51 & 26.67 \\
DeepSeek-R1-Distill-Qwen-7B~\cite{deepseek2024r1}$^{\textcolor{green}{\star}}$ & 53.09 & 32.89 & 35.56 & 34.57 & 39.47 & 28.89 \\
DS-R1-Qwen-32B~\cite{deepseek2024r1}$^{\textcolor{green}{\star}}$ & 64.20 & 43.42 & 42.22 & 51.85 & 52.63 & 37.78 \\
DeepSeek-R1~\cite{deepseek2024r1}$^{\textcolor{green}{\star}}$ & 74.19 & 55.13 & 44.44 & 74.19 & 71.79 & 55.56 \\
Llama3.2-3B-Instruct~\cite{grattafiori2024llama}$^{\textcolor{red}{\star}}$ & 47.31 & 26.92 & 26.67 & 44.09 & 43.59 & 46.67 \\
Mistral-Small-3.1-24B-Instruct~\cite{jung2010mistral}$^{\textcolor{red}{\star}}$ & 62.37 & 34.62 & 28.89 & 54.84 & 62.82 & 44.44 \\
Qwen2.5-Coder-3B~\cite{hui2024qwen2}$^{\textcolor{blue}{\star}}$ & 24.73 & 29.49 & 20.00 & 26.88 & 20.51 & 15.56 \\
Qwen2.5-3B-Instruct~\cite{yang2025qwen2}$^{\textcolor{red}{\star}}$ & 43.01 & 37.18 & 26.67 & 46.24 & 46.15 & 46.67 \\
Qwen2.5-72B-Ins~\cite{yang2025qwen2}$^{\textcolor{red}{\star}}$ & 67.74 & 38.46 & 24.44 & 50.54 & 47.43 & 51.11 \\
QwQ-32B~\cite{yang2025qwen3}$^{\textcolor{green}{\star}}$ & 62.37 & 34.62 & 33.33 & 53.76 & 57.69 & 37.78 \\
Qwen3-1.7B~\cite{yang2025qwen3}$^{\textcolor{red}{\star}}$ & 46.91 & 22.37 & 28.89 & 41.98 & 44.74 & 22.22 \\
Qwen3-8B~\cite{yang2025qwen3}$^{\textcolor{red}{\star}}$ & 70.96 & 43.59 & 22.22 & 44.09 & 46.15 & 42.22 \\
Qwen3-32B~\cite{yang2025qwen3}$^{\textcolor{red}{\star}}$ & 71.60 & 42.11 & 24.44 & 60.49 & 55.26 & 42.22 \\
Gemini-2.0-Flash~\cite{team2024gemini}$^{\textcolor{black}{\star}}$ & 77.78 & 40.79 & 31.11 & 62.96 & 55.26 & 51.11 \\
GPT-4o~\cite{hurst2024gpt}$^{\textcolor{black}{\star}}$ & 82.72 & 35.53 & 42.22 & 67.90 & 56.58 & 64.44 \\
Claude-3.7-Sonnet~\cite{anthropic2023claude2}$^{\textcolor{black}{\star}}$ & 80.25 & 47.37 & 33.33 & 77.78 & 65.79 & 71.11 \\
\bottomrule
\end{tabular}
\end{table*}

\begin{table*}[htbp]
\centering
\small
\caption{Performance on SVG editing dimension across different model types and difficulty levels. Results are reported using task-specific metrics (ACC, rMSE, MSE, RLD, CCR) for Bug Fixing, Style Editing and Code Optimization tasks, with models marked as specialized ({\textcolor{orange}{$\star$}}), reasoning ({\textcolor{green}{$\star$}}), code ({\textcolor{blue}{$\star$}}), open-source ({\textcolor{red}{$\star$}}), or proprietary ({\textcolor{black}{$\star$}}) variants.}
\label{tab:edition_task_full}
\begin{tabular}{l c c c c c c c }
\toprule
\multirow{2}{*}{Method} & \multicolumn{1}{c}{Bug Fixing} & \multicolumn{3}{c}{Style Editing} & \multicolumn{2}{c}{Code Optimization} \\
\cmidrule(lr){2-2} \cmidrule(lr){3-5} \cmidrule(lr){6-7} 
& ACC↑ & ACC↑ & rMSE↑ & RLD↓ & MSE↓ & CCR↑ \\
\midrule
\rowcolor{gray!20}
\multicolumn{7}{c}{\textbf{Easy}} \\
\hline
DeepSeek-R1-Distill-Qwen-1.5B~\cite{deepseek2024r1}$^{\textcolor{green}{\star}}$ & 3.85 & 56.06 & 28.29 & 234.41 & 11.50 & 13.04 \\
DeepSeek-R1-Distill-Qwen-7B~\cite{deepseek2024r1}$^{\textcolor{green}{\star}}$ & 3.80 & 69.69& 54.51&  136.09 &15.56 &18.35 \\
DS-R1-Dis-Qwen-32B~\cite{deepseek2024r1}$^{\textcolor{green}{\star}}$ & 62.63 & 84.81& 75.16 & 1.45 & 5.63 &23.16 \\
DeepSeek-R1~\cite{deepseek2024r1}$^{\textcolor{green}{\star}}$ & 71.00 & 84.62 & 73.66 & 18.21 & 1.01& 23.67 \\
Llama3.2-3B-Instruct~\cite{grattafiori2024llama}$^{\textcolor{red}{\star}}$ & 33.70 & 83.54 & 57.04 & 82.61 & 7.96  & 19.83 \\
Mistral-Small-3.1-24B-Instruct~\cite{jung2010mistral}$^{\textcolor{red}{\star}}$ & 54.00 & 74.36 & 62.39 & 25.84 & 12.05 & 23.51 \\
Qwen2.5-Coder-3B~\cite{hui2024qwen2}$^{\textcolor{blue}{\star}}$ & 9.43 & 56.82 & 39.55 & 136.92 & 17.29 & 20.90 \\
Qwen2.5-3B-Instruct~\cite{yang2025qwen2}$^{\textcolor{red}{\star}}$ & 36.46 & 70.51& 60.99 & 126.06 & 13.13 & 13.41 \\
Qwen2.5-72B-Ins~\cite{yang2025qwen2}$^{\textcolor{red}{\star}}$ & 71.00 & 75.95 & 64.25 & 2.08 & 2.98 & 20.57 \\
QwQ-32B~\cite{yang2025qwen3}$^{\textcolor{green}{\star}}$ & 71.43 &91.14 &81.70 &2.61 &3.20&20.20\\
Qwen3-1.7B~\cite{yang2025qwen3}$^{\textcolor{red}{\star}}$ & 22.34 & 74.36 & 67.22 & 33.40 & 11.43 & 35.79 \\
Qwen3-8B~\cite{yang2025qwen3}$^{\textcolor{red}{\star}}$ & 53.00 & 87.34& 80.40&221.88 &2.84 & 18.61\\
Qwen3-32B~\cite{yang2025qwen3}$^{\textcolor{red}{\star}}$ &56.12 &88.46& 82.63& 1.17 &2.55 & 17.96\\
Gemini-2.0-Flash~\cite{team2024gemini}$^{\textcolor{black}{\star}}$ & 69.00 & 86.07& 79.70&  4.78 & 0.72&  20.30 \\
GPT-4o~\cite{hurst2024gpt}$^{\textcolor{black}{\star}}$ & 74.00 & 78.48&65.81 & 46.53 & 1.30 & 10.57 \\
Claude-3.7-Sonnet~\cite{anthropic2023claude2}$^{\textcolor{black}{\star}}$ & 76.00 &79.75&67.17& 20.89 & 0.31& 16.81\\
\midrule
\rowcolor{gray!20}
\multicolumn{7}{c}{\textbf{Medium}} \\
\hline
DeepSeek-R1-Distill-Qwen-1.5B~\cite{deepseek2024r1}$^{\textcolor{green}{\star}}$ & 0.00 & 12.00 & 9.56 &  4757.67 & 14.73 & 47.38 \\
DeepSeek-R1-Distill-Qwen-7B~\cite{deepseek2024r1}$^{\textcolor{green}{\star}}$ & 0.00 &  36.36& 36.33 & 5811.5 & 19.05& 44.55 \\
DS-R1-Dis-Qwen-32B~\cite{deepseek2024r1}$^{\textcolor{green}{\star}}$ &46.00 & 56.41 & 53.11 &  140.68 &8.14 &33.10 \\
DeepSeek-R1~\cite{deepseek2024r1}$^{\textcolor{green}{\star}}$ & 63.83 & 62.16 & 53.93 & 1227.70 & 2.02&36.70 \\
Llama3.2-3B-Instruct~\cite{grattafiori2024llama}$^{\textcolor{blue}{\star}}$ & 4.60 & 50.67 & 37.93 & 442.66 & 5.33 & 15.53 \\
Mistral-Small-3.1-24B-Instruct~\cite{jung2010mistral}$^{\textcolor{red}{\star}}$ & 16.85 & 37.88 & 32.46 & 742.92 &  14.50 & 46.48 \\
Qwen2.5-Coder-3B~\cite{hui2024qwen2}$^{\textcolor{blue}{\star}}$ & 0.00 & 37.00 &29.96 & 3992.66 &  12.85 & 79.18 \\
Qwen2.5-3B-Instruct~\cite{yang2025qwen2}$^{\textcolor{red}{\star}}$ & 4.30 &34.33 & 27.11&1616.04 & 8.42 &19.81 \\
Qwen2.5-72B-Ins~\cite{yang2025qwen2}$^{\textcolor{red}{\star}}$ & 50.56 & 59.49& 51.90 & 136.00 &3.41 & 30.49\\
QwQ-32B~\cite{yang2025qwen3}$^{\textcolor{green}{\star}}$ & 46.00& 64.56& 61.27& 41.67 & 5.03&29.28 \\
Qwen3-1.7B~\cite{yang2025qwen3}$^{\textcolor{red}{\star}}$ & 1.22 & 33.90 & 25.55 &  1656.50 &10.20  & 34.87\\
Qwen3-8B~\cite{yang2025qwen3}$^{\textcolor{red}{\star}}$ &35.42 &62.34 &52.10 & 826.21 & 2.63&21.69 \\
Qwen3-32B~\cite{yang2025qwen3}$^{\textcolor{red}{\star}}$ & 46.47 &66.67 &62.54 & 17.13 &2.78 & 26.61\\
Gemini-2.0-Flash~\cite{team2024gemini}$^{\textcolor{black}{\star}}$ & 60.00 &65.82 &60.43 &113.44 & 1.59&27.49 \\
GPT-4o~\cite{hurst2024gpt}$^{\textcolor{black}{\star}}$ & 49.00 &50.63 & 42.64 & 840.73 &4.74 &43.01 \\
Claude-3.7-Sonnet~\cite{anthropic2023claude2}$^{\textcolor{black}{\star}}$ & 75.00 &59.74 &53.61&134.52 &0.86& 26.41 \\

\midrule
\rowcolor{gray!20}
\multicolumn{7}{c}{\textbf{Hard}} \\
\hline
DeepSeek-R1-Distill-Qwen-1.5B~\cite{deepseek2024r1}$^{\textcolor{green}{\star}}$ & 0.00 & 28.57 & 6933.75 & 352.85 & 12.38 & 57.41 \\
DeepSeek-R1-Distill-Qwen-7B~\cite{deepseek2024r1}$^{\textcolor{green}{\star}}$ &0.00 & 42.86 & 42.85 &  11559.67 & 8.80 & 61.35 \\
DS-R1-Dis-Qwen-32B~\cite{deepseek2024r1}$^{\textcolor{green}{\star}}$ & 34.02 & 55.26& 41.68 & 226.50 & 5.89 &36.41 \\
Deepseek-R1~\cite{deepseek2024r1}$^{\textcolor{green}{\star}}$ & 61.80 & 51.56 & 41.83 & 2610.48 &2.41 &39.95 \\
Llama3.2-3B-Instruct~\cite{grattafiori2024llama}$^{\textcolor{red}{\star}}$ & 1.37 & 40.35 & 24.03 & 3631.91 & 5.34 & 9.52 \\
mistral-small-3.1-24b-instruct~\cite{jung2010mistral}$^{\textcolor{red}{\star}}$ &  3.95 &25.86 & 20.25 & 2247.67 & 11.99 &  55.31\\
Qwen2.5-coder-3B~\cite{hui2024qwen2}$^{\textcolor{blue}{\star}}$ & 0.00 & 36.36 & 32.81 & 7347.00 & 12.39& 82.32 \\
Qwen2.5-3B-Instruct~\cite{yang2025qwen2}$^{\textcolor{red}{\star}}$ & 0.00 &  24.53& 15.90 & 4122.69 & 4.94& 20.76\\
Qwen2.5-72B-Ins~\cite{yang2025qwen2}$^{\textcolor{red}{\star}}$ & 40.00 & 54.67 & 43.80 & 722.00 & 3.24&36.40 \\
QwQ-32B~\cite{yang2025qwen3}$^{\textcolor{green}{\star}}$ &10.42 &50.00 & 44.04& 533.85&  5.82&12.98 \\
Qwen3-1.7B~\cite{yang2025qwen3}$^{\textcolor{red}{\star}}$ & 0.00& 26.42 &21.94 & 3650.71 & 5.60 & 32.42 \\
Qwen3-8B~\cite{yang2025qwen3}$^{\textcolor{red}{\star}}$ & 25.30& 60.81& 52.21& 935.13 &2.15 &23.68 \\
Qwen3-32B~\cite{yang2025qwen3}$^{\textcolor{red}{\star}}$ &40.86 & 55.26&48.06 & 58.79 &1.50 &26.75 \\
Gemini-2.0-Flash~\cite{team2024gemini}$^{\textcolor{black}{\star}}$ & 53.95& 57.38&42.34 & 628.14 &2.62 &29.35 \\
GPT-4o~\cite{hurst2024gpt}$^{\textcolor{black}{\star}}$ & 51.02 & 42.67& 35.43 & 843.81 & 5.14& 46.91\\
Claude-3.7-Sonnet~\cite{anthropic2023claude2}$^{\textcolor{black}{\star}}$ & 69.00 &52.56&40.75& 27.00 &0.88 &28.27 \\

\bottomrule
\end{tabular}
\end{table*}

\begin{table*}[htbp]
\centering
\small
\caption{Performance on SVG generation dimension across different model types and difficulty levels. Results are reported using task-specific metrics (CLIP, AES, HPS, rCLIP, PSS, Cart., Pixel, Line, 3D) for Text-based Generation and Style Transfer tasks, with models marked as  specialized ({\textcolor{orange}{$\star$}}), reasoning ({\textcolor{green}{$\star$}}), code ({\textcolor{blue}{$\star$}}), open-source ({\textcolor{red}{$\star$}}), or proprietary ({\textcolor{black}{$\star$}}) variants.}
\label{tab:generation_full}
\begin{tabular}{l c c c c c c c c c}
\toprule
\multirow{2}{*}{Method} & \multicolumn{5}{c}{Text-based Generation} & \multicolumn{4}{c}{Style Transfer} \\
\cmidrule(lr){2-6} \cmidrule(lr){7-10}
& CLIP↑ & AES↑ & HPS↑ & rCLIP↑ & PSS↑ & Cart. & Pixel & Line & 3D \\
\midrule
\rowcolor{gray!20}
\multicolumn{10}{c}{\textbf{Easy}} \\
\hline
Iconshop~\cite{wu2023iconshop}$^{\textcolor{orange}{\star}}$ & 22.74 & 3.56 & 17.99 & 86.22 & 5.26 & - & - & - & - \\
LLM4SVG(GPT-2 XL)~\cite{xing2024empowering}$^{\textcolor{orange}{\star}}$ & 18.09 & 3.63 & 16.76 & 75.78 & 3.01 & - & - & - & - \\
DeepSeek-R1-Distill-Qwen-1.5B~\cite{deepseek2024r1}$^{\textcolor{green}{\star}}$ & 18.43 & 3.55 & 16.51 & 73.51 & 0.79 & 1.27 & 1.25 & 0.36 & 1.30 \\
DeepSeek-R1-Distill-Qwen-7B~\cite{deepseek2024r1}$^{\textcolor{green}{\star}}$ & 19.27 & 3.50 & 16.71 & 76.76 & 0.93 & 0.77 & 0.65 & 1.60 & 0.90 \\
DS-R1-Qwen-32B~\cite{deepseek2024r1}$^{\textcolor{green}{\star}}$ & 22.04 & 3.47 & 17.42 & 84.09 & 10.47 & 3.14 & 1.55 & 3.04 & 2.73 \\
DeepSeek-R1~\cite{deepseek2024r1}$^{\textcolor{green}{\star}}$ & 24.34 & 3.61 & 20.37 & 91.39 & 18.07 & 3.13 & 1.70 & 2.68 & 2.87 \\
Mistral-Small-3.1-24B-Instruct~\cite{jung2010mistral}$^{\textcolor{red}{\star}}$ & 21.15 & 3.25 & 16.64 & 81.81 & 8.31 & 2.07 & 1.65 & 2.28 & 1.30 \\
Qwen2.5-Coder-3B~\cite{hui2024qwen2}$^{\textcolor{blue}{\star}}$ & 19.23 & 3.22 & 14.97 & 74.08 & 1.32 & - & - & - & - \\
Qwen2.5-3B-Instruct~\cite{yang2025qwen2}$^{\textcolor{red}{\star}}$ & 20.03 & 3.19 & 15.81 & 78.25 & 2.33 & 1.70 & 1.87 & 1.68 & 2.17 \\
Qwen2.5-72B-Ins~\cite{yang2025qwen2}$^{\textcolor{red}{\star}}$ & 20.86 & 3.59 & 17.38 & 91.18 & 15.77 & 2.83 & 1.85 & 2.72 & 2.90 \\
QwQ-32B~\cite{yang2025qwen3}$^{\textcolor{green}{\star}}$ & 23.01 & 3.51 & 19.19 & 89.30 & 16.78 & 3.56 & 2.24 & 3.30 & 3.12 \\
Qwen3-1.7B~\cite{yang2025qwen3}$^{\textcolor{red}{\star}}$ & 20.37 & 3.50 & 16.92 & 79.80 & 10.22 & 1.64 & 2.35 & 3.00 & 1.97 \\
Qwen3-8B~\cite{yang2025qwen3}$^{\textcolor{red}{\star}}$ & 22.39 & 3.48 & 18.15 & 85.53 & 12.73 & - & - & - & - \\
Qwen3-32B~\cite{yang2025qwen3}$^{\textcolor{red}{\star}}$ & 23.81 & 3.48 & 19.39 & 89.13 & 12.62 & 2.93 & 2.10 & 2.92 & 2.80 \\
Gemini-2.0-Flash~\cite{team2024gemini}$^{\textcolor{black}{\star}}$ & 24.20 & 3.70 & 19.82 & 90.96 & 15.94 & 3.17 & 2.32 & 3.16 & 3.25 \\
GPT-4o~\cite{hurst2024gpt}$^{\textcolor{black}{\star}}$ & 24.97 & 3.63 & 20.35 & 90.95 & 19.72 & 3.27 & 1.95 & 2.88 & 2.53 \\
Claude-3.7-Sonnet~\cite{anthropic2023claude2}$^{\textcolor{black}{\star}}$ & 25.11 & 3.96 & 21.35 & 92.90 & 16.69 & 3.68 & 2.00 & 2.08 & 2.93 \\
\midrule
\rowcolor{gray!20}
\multicolumn{10}{c}{\textbf{Medium}} \\
\hline
Iconshop~\cite{wu2023iconshop}$^{\textcolor{orange}{\star}}$ & 20.76 & 3.46 & 14.68 & 77.35 & 3.24 & - & - & - & - \\
LLM4SVG(GPT-2 XL)~\cite{xing2024empowering}$^{\textcolor{orange}{\star}}$ & 17.98 & 3.55 & 14.88 & 68.84 & 2.70 & - & - & - & - \\
DeepSeek-R1-Distill-Qwen-1.5B~\cite{deepseek2024r1}$^{\textcolor{green}{\star}}$ & 17.23 & 3.49 & 14.99 & 66.43 & 0.23 & 1.13 & 0.53 & 0.00 & 0.00 \\
DeepSeek-R1-Distill-Qwen-7B~\cite{deepseek2024r1}$^{\textcolor{green}{\star}}$ & 18.21 & 3.53 & 15.00 & 69.20 & 0.41 & 0.00 & 0.65 & 0.62 & 0.52 \\
DS-R1-Qwen-32B~\cite{deepseek2024r1}$^{\textcolor{green}{\star}}$ & 20.04 & 3.52 & 15.76 & 75.52 & 7.55 & 1.27 & 1.43 & 2.38 & 3.16 \\
DeepSeek-R1~\cite{deepseek2024r1}$^{\textcolor{green}{\star}}$ & 22.54 & 3.55 & 17.46 & 82.99 & 14.78 & 2.40 & 1.82 & 2.12 & 2.44 \\
Mistral-Small-3.1-24B-Instruct~\cite{jung2010mistral}$^{\textcolor{red}{\star}}$ & 20.09 & 3.32 & 15.34 & 75.40 & 5.39 & 1.80 & 0.68 & 1.38 & 0.00 \\
Qwen2.5-Coder-3B~\cite{hui2024qwen2}$^{\textcolor{blue}{\star}}$ & 16.75 & 3.38 & 13.48 & 63.74 & 1.89 & - & - & - & - \\
Qwen2.5-3B-Instruct~\cite{yang2025qwen2}$^{\textcolor{red}{\star}}$ & 18.08 & 3.32 & 13.40 & 69.46 & 1.47 & 1.00 & 1.18 & 1.48 & 1.12 \\
Qwen2.5-72B-Ins~\cite{yang2025qwen2}$^{\textcolor{red}{\star}}$ & 20.40 & 3.50 & 16.35 & 79.91 & 13.11 & 3.07 & 2.13 & 2.12 & 3.52 \\
QwQ-32B~\cite{yang2025qwen3}$^{\textcolor{green}{\star}}$ & 21.39 & 3.57 & 16.73 & 77.87 & 17.56 & 3.00 & 1.47 & 2.31 & 3.48 \\
Qwen3-1.7B~\cite{yang2025qwen3}$^{\textcolor{red}{\star}}$ & 19.26 & 3.59 & 16.00 & 70.66 & 11.97 & 1.87 & 2.14 & 1.89 & 1.84 \\
Qwen3-8B~\cite{yang2025qwen3}$^{\textcolor{red}{\star}}$ & 21.65 & 3.45 & 16.70 & 80.37 & 11.26 & - & - & - & - \\
Qwen3-32B~\cite{yang2025qwen3}$^{\textcolor{red}{\star}}$ & 21.80 & 3.51 & 17.17 & 80.88 & 14.38 & 3.33 & 1.67 & 1.56 & 2.76 \\
Gemini-2.0-Flash~\cite{team2024gemini}$^{\textcolor{black}{\star}}$ & 21.77 & 3.65 & 17.00 & 80.36 & 13.70 & 3.13 & 1.87 & 2.13 & 3.00 \\
GPT-4o~\cite{hurst2024gpt}$^{\textcolor{black}{\star}}$ & 23.03 & 3.49 & 17.51 & 84.72 & 11.97 & 1.67 & 1.69 & 2.00 & 1.56 \\
Claude-3.7-Sonnet~\cite{anthropic2023claude2}$^{\textcolor{black}{\star}}$ & 24.18 & 3.78 & 85.71 & 87.62 & 16.60 & 3.67 & 2.61 & 1.94 & 3.28 \\
\midrule
\rowcolor{gray!20}
\multicolumn{10}{c}{\textbf{Hard}} \\
\hline
Iconshop~\cite{wu2023iconshop}$^{\textcolor{orange}{\star}}$ & 19.34 & 3.48 & 12.95 & 72.55 & 2.12 & - & - & - & - \\
LLM4SVG(GPT-2 XL)~\cite{xing2024empowering}$^{\textcolor{orange}{\star}}$ & 16.19 & 3.61 & 14.62 & 62.98 & 0.02 & - & - & - & - \\
DeepSeek-R1-Distill-Qwen-1.5B~\cite{deepseek2024r1}$^{\textcolor{green}{\star}}$ & 16.19 & 3.46 & 14.01 & 61.25 & 0.12 & 0.00 & 0.08 & 0.32 & 0.00 \\
DeepSeek-R1-Distill-Qwen-7B~\cite{deepseek2024r1}$^{\textcolor{green}{\star}}$ & 17.51 & 3.48 & 14.10 & 65.33 & 0.25 & 0.24 & 0.35 & 1.04 & 0.00 \\
DS-R1-Qwen-32B~\cite{deepseek2024r1}$^{\textcolor{green}{\star}}$ & 19.22 & 3.63 & 14.56 & 72.71 & 6.56 & 2.20 & 1.50 & 2.08 & 2.40 \\
DeepSeek-R1~\cite{deepseek2024r1}$^{\textcolor{green}{\star}}$ & 22.73 & 3.63 & 16.86 & 83.06 & 10.07 & 2.16 & 1.95 & 2.00 & 2.13 \\
Mistral-Small-3.1-24B-Instruct~\cite{jung2010mistral}$^{\textcolor{red}{\star}}$ & 18.61 & 3.33 & 14.35 & 70.60 & 4.11 & 2.07 & 1.65 & 2.28 & 1.30 \\
Qwen2.5-Coder-3B~\cite{hui2024qwen2}$^{\textcolor{blue}{\star}}$ & 16.07 & 3.35 & 12.50 & 59.91 & 1.31 & - & - & - & - \\
Qwen2.5-3B-Instruct~\cite{yang2025qwen2}$^{\textcolor{red}{\star}}$ & 17.38 & 3.30 & 13.31 & 65.59 & 2.08 & 0.96 & 1.15 & 1.24 & 1.20 \\
Qwen2.5-72B-Ins~\cite{yang2025qwen2}$^{\textcolor{red}{\star}}$ & 20.08 & 3.65 & 15.51 & 75.22 & 9.57 & 2.64 & 2.00 & 1.96 & 3.20 \\
QwQ-32B~\cite{yang2025qwen3}$^{\textcolor{green}{\star}}$ & 22.03 & 3.51 & 16.36 & 82.24 & 8.45 & 1.92 & 1.40 & 1.92 & 2.40 \\
Qwen3-1.7B~\cite{yang2025qwen3}$^{\textcolor{red}{\star}}$ & 18.37 & 3.50 & 14.52 & 68.91 & 4.46 & 1.32 & 1.30 & 2.24 & 1.20 \\
Qwen3-8B~\cite{yang2025qwen3}$^{\textcolor{red}{\star}}$ & 20.60 & 3.43 & 15.34 & 77.47 & 8.64 & - & - & - & - \\
Qwen3-32B~\cite{yang2025qwen3}$^{\textcolor{red}{\star}}$ & 22.06 & 3.52 & 16.02 & 81.84 & 9.88 & 2.96 & 1.27 & 1.60 & 3.33 \\
Gemini-2.0-Flash~\cite{team2024gemini}$^{\textcolor{black}{\star}}$ & 20.95 & 3.73 & 16.01 & 78.61 & 9.89 & 1.32 & 1.39 & 1.08 & 1.26 \\
GPT-4o~\cite{hurst2024gpt}$^{\textcolor{black}{\star}}$ & 22.57 & 3.64 & 16.69 & 82.81 & 10.39 & 1.84 & 1.92 & 1.96 & 2.73 \\
Claude-3.7-Sonnet~\cite{anthropic2023claude2}$^{\textcolor{black}{\star}}$ & 24.54 & 3.96 & 18.74 & 87.97 & 10.19 & 2.64 & 2.36 & 1.80 & 3.33 \\
\bottomrule
\end{tabular}
\end{table*}

\begin{table*}[htbp]
\centering
\small
\caption{Performance on SVG generation dimension across different model types and difficulty levels. Results are reported using task-specific metrics (CP, DF, SC, CH, CB) for Style Transfer tasks, with models marked as reasoning ({\textcolor{green}{$\star$}}), open-source ({\textcolor{red}{$\star$}}), or proprietary ({\textcolor{black}{$\star$}}) variants.}
\label{tab:style_transfer_task}
\begin{tabular}{l C{0.3cm} C{0.3cm}C{0.3cm}C{0.3cm}C{0.3cm}C{0.3cm}C{0.3cm}C{0.3cm}C{0.3cm}C{0.3cm}C{0.3cm}C{0.3cm}C{0.3cm}C{0.3cm}C{0.3cm}C{0.3cm}C{0.3cm}C{0.3cm}C{0.3cm}C{0.3cm}}
\toprule
\multirow{2}{*}{Method} & \multicolumn{5}{c}{Cartoon Style} & \multicolumn{5}{c}{Pixel art} & \multicolumn{5}{c}{Line art} & \multicolumn{5}{c}{3D style}\\
\cmidrule(lr){2-6} \cmidrule(lr){7-11} \cmidrule(lr){12-16} \cmidrule(lr){17-21}
& CP & DF & SC & CH & CB & CP & DF & SC & CH & CB & CP & DF & SC & CH & CB & CP & DF & SC & CH & CB \\
\midrule
\rowcolor{gray!20}
\multicolumn{21}{c}{\textbf{Easy}} \\
\hline
DeepSeek-R1-Distill-Qwen-1.5B~\cite{deepseek2024r1}$^{\textcolor{green}{\star}}$ & 1.17 & 0.67 & 0.67 & 1.17 & 2.67 & 0.75 & 0.75 & 0.75 & 1.25 & 2.75 & 0.40 & 0.20 & 0.40 & 0.40 & 0.40 & 1.17 & 1.00 & 0.67 & 1.83 & 1.83 \\
DeepSeek-R1-Distill-Qwen-7B~\cite{deepseek2024r1}$^{\textcolor{green}{\star}}$ & 0.33 & 0.33 & 0.83 & 0.67 & 1.67 & 0.50 & 0.50 & 0.50 & 0.75 & 1.00 & 1.20 & 1.20 & 1.60 & 1.60 & 2.40 & 0.67 & 0.67 & 0.67 & 1.17 & 1.33 \\
DS-R1-Qwen-32B~\cite{deepseek2024r1}$^{\textcolor{green}{\star}}$ & 3.17 & 2.67 & 2.67 & 2.17 & 5.00 & 1.25 & 1.25 & 1.25 & 1.75 & 2.25 & 2.80 & 2.60 & 3.00 & 2.40 & 4.40 & 2.83 & 2.00 & 2.33 & 3.00 & 3.50 \\
DeepSeek-R1~\cite{deepseek2024r1}$^{\textcolor{green}{\star}}$ & 3.67 & 2.00 & 2.83 & 2.67 & 4.50 & 1.50 & 1.25 & 1.25 & 2.00 & 2.50 & 2.20 & 2.20 & 3.00 & 1.80 & 4.20 & 3.17 & 2.33 & 2.33 & 3.00 & 3.50 \\
Llama3.2-3B-Instruct~\cite{grattafiori2024llama}$^{\textcolor{red}{\star}}$ & 0.83 & 1.00 & 0.83 & 1.33 & 1.83 & 1.75 & 1.75 & 1.50 & 2.25 & 2.75 & 1.00 & 0.80 & 0.40 & 0.60 & 1.40 & 0.83 & 0.67 & 0.67 & 1.17 & 1.33 \\
Mistral-Small-3.1-24B-Instruct~\cite{jung2010mistral}$^{\textcolor{red}{\star}}$ & 2.00 & 1.67 & 1.83 & 1.50 & 3.33 & 1.25 & 1.25 & 1.00 & 2.00 & 2.75 & 1.60 & 2.40 & 2.00 & 1.40 & 4.00 & 1.50 & 1.17 & 0.83 & 1.00 & 2.00 \\
Qwen2.5-3B-Instruct~\cite{yang2025qwen2}$^{\textcolor{red}{\star}}$ & 1.50 & 1.33 & 1.17 & 1.50 & 3.00 & 1.50 & 1.00 & 1.17 & 2.17 & 3.50 & 1.60 & 1.00 & 1.20 & 2.20 & 2.40 & 2.33 & 1.67 & 1.00 & 2.83 & 3.00 \\
Qwen2.5-72B-Ins~\cite{yang2025qwen2}$^{\textcolor{red}{\star}}$ & 3.17 & 2.00 & 1.83 & 2.67 & 4.50 & 1.25 & 1.00 & 1.50 & 2.50 & 3.00 & 2.20 & 2.60 & 2.80 & 1.40 & 4.60 & 3.50 & 2.33 & 2.17 & 2.83 & 3.67 \\
QwQ-32B~\cite{yang2025qwen3}$^{\textcolor{green}{\star}}$ & 3.80 & 2.80 & 3.80 & 2.80 & 4.60 & 1.60 & 1.80 & 1.40 & 2.80 & 3.60 & 3.00 & 2.50 & 3.75 & 2.75 & 4.50 & 4.00 & 2.00 & 2.60 & 3.00 & 4.00 \\
Qwen3-1.7B~\cite{yang2025qwen3}$^{\textcolor{red}{\star}}$ & 1.17 & 1.00 & 1.17 & 1.67 & 3.17 & 2.00 & 2.00 & 1.75 & 2.50 & 3.50 & 3.00 & 3.00 & 2.80 & 1.80 & 4.40 & 2.67 & 1.50 & 1.00 & 2.17 & 2.50 \\
Qwen3-4B~\cite{yang2025qwen3}$^{\textcolor{red}{\star}}$ & 3.17 & 2.00 & 2.67 & 2.50 & 4.33 & 2.75 & 3.00 & 2.75 & 3.25 & 3.50 & 1.80 & 2.20 & 2.40 & 0.80 & 3.60 & 3.50 & 2.67 & 2.33 & 3.00 & 3.83 \\
Qwen3-32B~\cite{yang2025qwen3}$^{\textcolor{red}{\star}}$ & 3.17 & 2.33 & 2.67 & 2.50 & 4.00 & 1.50 & 1.50 & 1.75 & 2.75 & 3.00 & 3.60 & 2.80 & 1.80 & 1.80 & 4.60 & 3.00 & 2.33 & 2.67 & 2.83 & 3.17 \\
Gemini-2.0-Flash~\cite{team2024gemini}$^{\textcolor{black}{\star}}$ & 3.17 & 2.67 & 3.17 & 2.67 & 4.17 & 1.75 & 1.50 & 1.75 & 2.25 & 4.33 & 3.60 & 3.00 & 2.20 & 2.20 & 4.80 & 3.67 & 2.67 & 3.40 & 3.00 & 3.50 \\
GPT-4o~\cite{hurst2024gpt}$^{\textcolor{black}{\star}}$ & 3.67 & 2.67 & 2.67 & 2.83 & 4.50 & 1.50 & 1.50 & 1.50 & 2.25 & 3.00 & 3.20 & 2.00 & 3.40 & 1.80 & 4.00 & 3.00 & 2.00 & 1.83 & 2.50 & 3.33 \\
Claude-3.7-Sonnet~\cite{anthropic2023claude2}$^{\textcolor{black}{\star}}$ & 4.00 & 2.75 & 3.83 & 3.00 & 4.83 & 1.50 & 1.50 & 1.75 & 2.25 & 3.00 & 2.00 & 1.40 & 1.40 & 1.20 & 4.40 & 3.33 & 2.50 & 2.33 & 2.67 & 3.83 \\
\midrule
\rowcolor{gray!20}
\multicolumn{21}{c}{\textbf{Medium}} \\
\hline
DeepSeek-R1-Distill-Qwen-1.5B~\cite{deepseek2024r1}$^{\textcolor{green}{\star}}$ & 0.67 & 0.67 & 0.67 & 1.33 & 2.33 & 0.42 & 0.58 & 0.33 & 0.50 & 0.83 & 0.00 & 0.00 & 0.00 & 0.00 & 0.00 & 0.00 & 0.00 & 0.00 & 0.00 & 0.00 \\
DeepSeek-R1-Distill-Qwen-7B~\cite{deepseek2024r1}$^{\textcolor{green}{\star}}$ & 0.00 & 0.00 & 0.00 & 0.00 & 0.00 & 0.58 & 0.50 & 0.42 & 0.75 & 1.00 & 0.50 & 0.50 & 0.60 & 0.60 & 0.90 & 0.40 & 0.40 & 0.40 & 0.60 & 0.80 \\
DS-R1-Qwen-32B~\cite{deepseek2024r1}$^{\textcolor{green}{\star}}$ & 1.00 & 1.33 & 1.00 & 1.33 & 1.67 & 1.25 & 1.17 & 1.17 & 1.58 & 2.00 & 2.50 & 1.80 & 1.60 & 1.70 & 4.30 & 3.20 & 3.00 & 3.20 & 2.80 & 3.60 \\
DeepSeek-R1~\cite{deepseek2024r1}$^{\textcolor{green}{\star}}$ & 2.00 & 1.67 & 2.00 & 2.67 & 3.67 & 1.17 & 1.42 & 1.83 & 2.00 & 2.67 & 1.90 & 1.90 & 1.80 & 1.60 & 3.40 & 2.40 & 2.80 & 2.00 & 2.60 & 2.40 \\
Llama3.2-3B-Instruct~\cite{grattafiori2024llama}$^{\textcolor{red}{\star}}$ & 0.67 & 0.67 & 0.67 & 1.00 & 0.67 & 0.75 & 0.75 & 0.64 & 1.17 & 1.25 & 1.00 & 0.90 & 0.17 & 0.70 & 1.50 & 1.40 & 1.20 & 1.40 & 1.20 & 1.60 \\
Mistral-Small-3.1-24B-Instruct~\cite{jung2010mistral}$^{\textcolor{red}{\star}}$ & 1.67 & 1.33 & 1.67 & 1.67 & 2.67 & 0.42 & 0.42 & 0.50 & 1.00 & 1.08 & 1.10 & 1.10 & 1.50 & 0.90 & 2.30 & 0.00 & 0.00 & 0.00 & 0.00 & 0.00 \\
Qwen2.5-3B-Instruct~\cite{yang2025qwen2}$^{\textcolor{red}{\star}}$ & 0.75 & 0.75 & 1.00 & 1.25 & 1.25 & 1.00 & 0.92 & 0.83 & 1.33 & 1.82 & 1.30 & 1.20 & 1.10 & 1.80 & 2.00 & 0.80 & 0.60 & 0.80 & 1.80 & 1.60 \\
Qwen2.5-72B-Ins~\cite{yang2025qwen2}$^{\textcolor{red}{\star}}$ & 3.00 & 2.67 & 2.67 & 3.00 & 4.00 & 2.00 & 1.75 & 1.50 & 2.25 & 3.17 & 1.90 & 1.60 & 2.30 & 1.20 & 3.60 & 3.60 & 2.80 & 3.60 & 3.40 & 4.20 \\
QwQ-32B~\cite{yang2025qwen3}$^{\textcolor{green}{\star}}$ & 2.67 & 3.00 & 3.33 & 2.67 & 3.33 & 1.36 & 1.21 & 1.21 & 1.43 & 2.14 & 2.09 & 1.91 & 2.45 & 1.64 & 3.45 & 3.80 & 3.00 & 3.60 & 3.40 & 3.60 \\
Qwen3-1.7B~\cite{yang2025qwen3}$^{\textcolor{red}{\star}}$ & 2.00 & 2.00 & 0.67 & 2.00 & 2.67 & 1.91 & 2.20 & 1.83 & 2.25 & 2.50 & 2.20 & 1.44 & 1.30 & 1.40 & 3.10 & 2.20 & 1.80 & 1.20 & 1.80 & 2.20 \\
Qwen3-4B~\cite{yang2025qwen3}$^{\textcolor{red}{\star}}$ & 3.33 & 3.00 & 2.33 & 2.33 & 3.33 & 3.00 & 2.67 & 2.92 & 2.58 & 3.17 & 2.20 & 2.00 & 2.40 & 1.20 & 3.70 & 4.00 & 2.60 & 2.60 & 3.60 & 3.60 \\
Qwen3-32B~\cite{yang2025qwen3}$^{\textcolor{red}{\star}}$ & 3.33 & 3.00 & 2.67 & 3.00 & 4.67 & 1.00 & 1.25 & 1.50 & 2.00 & 2.58 & 1.60 & 1.30 & 1.10 & 1.20 & 2.58 & 2.60 & 2.40 & 2.80 & 2.60 & 3.40 \\
Gemini-2.0-Flash~\cite{team2024gemini}$^{\textcolor{black}{\star}}$ & 3.00 & 3.00 & 3.33 & 3.00 & 3.33 & 1.64 & 1.42 & 1.58 & 2.25 & 2.45 & 1.67 & 1.90 & 2.10 & 1.30 & 3.70 & 3.00 & 2.80 & 3.20 & 2.80 & 3.20 \\
GPT-4o~\cite{hurst2024gpt}$^{\textcolor{black}{\star}}$ & 1.67 & 2.00 & 1.00 & 1.33 & 2.33 & 1.45 & 1.55 & 1.45 & 1.91 & 2.09 & 2.00 & 1.30 & 1.30 & 1.40 & 4.00 & 1.60 & 1.60 & 1.40 & 1.60 & 1.60 \\
Claude-3.7-Sonnet~\cite{anthropic2023claude2}$^{\textcolor{black}{\star}}$ & 3.00 & 3.00 & 4.00 & 3.33 & 5.00 & 2.50 & 2.33 & 2.58 & 2.83 & 2.83 & 1.60 & 1.40 & 1.30 & 1.50 & 3.90 & 3.60 & 3.00 & 3.00 & 2.80 & 4.00 \\
\midrule
\rowcolor{gray!20}
\multicolumn{21}{c}{\textbf{Hard}} \\
\hline
DeepSeek-R1-Distill-Qwen-1.5B~\cite{deepseek2024r1}$^{\textcolor{green}{\star}}$ & 0.00 & 0.00 & 0.00 & 0.00 & 0.00 & 0.06 & 0.06 & 0.06 & 0.12 & 0.12 & 0.20 & 0.20 & 0.20 & 0.20 & 0.80 & 0.00 & 0.00 & 0.00 & 0.00 & 0.00 \\
DeepSeek-R1-Distill-Qwen-7B~\cite{deepseek2024r1}$^{\textcolor{green}{\star}}$ & 0.20 & 0.20 & 0.20 & 0.20 & 0.40 & 0.19 & 0.19 & 0.25 & 0.50 & 0.62 & 0.60 & 1.00 & 1.40 & 0.60 & 1.60 & 0.00 & 0.00 & 0.00 & 0.00 & 0.00 \\
DS-R1-Qwen-32B~\cite{deepseek2024r1}$^{\textcolor{green}{\star}}$ & 2.20 & 1.80 & 1.60 & 2.20 & 3.20 & 1.25 & 1.19 & 1.44 & 1.69 & 1.94 & 1.00 & 1.40 & 2.60 & 1.20 & 4.20 & 2.67 & 2.00 & 1.67 & 2.67 & 3.00 \\
DeepSeek-R1~\cite{deepseek2024r1}$^{\textcolor{green}{\star}}$ & 1.40 & 1.60 & 1.80 & 2.60 & 3.40 & 1.60 & 1.25 & 1.88 & 2.19 & 2.81 & 1.00 & 1.40 & 3.20 & 1.20 & 3.20 & 2.00 & 1.67 & 2.00 & 2.67 & 2.33\\
Llama3.2-3B-Instruct~\cite{grattafiori2024llama}$^{\textcolor{red}{\star}}$ & 0.83 & 1.00 & 0.83 & 1.33 & 1.83 & 1.75 & 1.75 & 1.50 & 2.25 & 2.75 & 1.00 & 0.80 & 0.40 & 0.60 & 1.40 & 0.83 & 0.67 & 0.67 & 1.17 & 1.33 \\
Mistral-Small-3.1-24B-Instruct~\cite{jung2010mistral}$^{\textcolor{red}{\star}}$ & 2.0 & 1.67 & 1.83 & 1.5 & 3.33 & 1.25 & 1.25 & 1.0 & 2.0 & 2.75 & 1.6 & 2.4 & 2.0 & 1.4 & 4.0 & 1.5 & 1.17 & 0.83 & 1.0 & 2.0 \\
Qwen2.5-3B-Instruct~\cite{yang2025qwen2}$^{\textcolor{red}{\star}}$ & 0.80 & 0.80 & 0.80 & 0.80 & 1.60 & 0.79 & 0.79 & 0.95 & 1.32 & 1.89 & 1.00 & 1.00 & 1.00 & 1.60 & 1.60 & 1.00 & 1.00 & 1.00 & 1.33 & 1.67 \\
Qwen2.5-72B-Ins~\cite{yang2025qwen2}$^{\textcolor{red}{\star}}$ & 3.40 & 2.60 & 1.60 & 2.00 & 3.60 & 1.94 & 1.69 & 1.62 & 2.12 & 2.62 & 1.20 & 1.40 & 2.40 & 1.00 & 3.80 & 3.33 & 3.33 & 2.00 & 3.33 & 4.00 \\
QwQ-32B~\cite{yang2025qwen3}$^{\textcolor{green}{\star}}$ & 1.80 & 1.60 & 1.60 & 1.80 & 2.80 & 0.94 & 1.00 & 1.31 & 1.62 & 2.12 & 1.40 & 1.60 & 2.20 & 1.80 & 2.60 & 2.33 & 2.00 & 2.67 & 2.33 & 2.67 \\
Qwen3-1.7B~\cite{yang2025qwen3}$^{\textcolor{red}{\star}}$ & 1.00 & 1.00 & 1.00 & 1.40 & 2.20 & 1.27 & 1.12 & 1.12 & 1.19 & 1.81 & 2.20 & 1.80 & 2.80 & 1.60 & 2.80 & 0.67 & 0.67 & 1.00 & 1.67 & 2.00 \\
Qwen3-4B~\cite{yang2025qwen3}$^{\textcolor{red}{\star}}$ & 2.60 & 2.20 & 2.60 & 2.20 & 4.20 & 3.06 & 2.44 & 2.81 & 3.00 & 3.69 & 1.80 & 1.80 & 3.40 & 1.60 & 4.20 & 2.67 & 2.33 & 2.67 & 2.00 & 3.33 \\
Qwen3-32B~\cite{yang2025qwen3}$^{\textcolor{red}{\star}}$ & 3.20 & 2.60 & 2.00 & 2.80 & 4.20 & 1.12 & 1.00 & 1.06 & 1.25 & 1.94 & 1.00 & 1.00 & 1.60 & 1.00 & 3.40 & 4.00 & 3.00 & 3.33 & 2.67 & 3.67 \\
Gemini-2.0-Flash~\cite{team2024gemini}$^{\textcolor{black}{\star}}$ & 1.20 & 1.00 & 1.20 & 1.40 & 1.80 & 1.00 & 0.94 & 1.25 & 1.56 & 2.19 & 0.60 & 1.00 & 1.20 & 0.60 & 2.00 & 1.33 & 1.00 & 1.33 & 1.33 & 1.33 \\
GPT-4o~\cite{hurst2024gpt}$^{\textcolor{black}{\star}}$ & 1.40 & 1.20 & 1.40 & 1.60 & 3.60 & 1.69 & 1.44 & 1.81 & 2.12 & 2.56 & 1.00 & 1.40 & 2.20 & 1.40& 3.80 & 3.00 & 3.00 & 2.00 & 2.33 & 3.33 \\
Claude-3.7-Sonnet~\cite{anthropic2023claude2}$^{\textcolor{black}{\star}}$ & 2.60 & 2.40 & 2.20 & 2.20 & 3.80 & 2.25 & 2.06 & 2.06 & 2.44 & 3.00 & 1.00 & 1.40 & 1.60 & 1.20 & 3.80 & 3.67 & 3.00 & 3.00 & 3.00 & 4.00 \\
\bottomrule
\end{tabular}
\end{table*}

\section{Path-Structure Similarity Score}
\label{app:evaluation metrics}
We propose a novel evaluation metric to evaluate the quality of SVG image generation, termed the Path-Structure Similarity Score (PSS). Unlike conventional image quality metrics such as MSE, SSIM~\cite{wang2004image}, or FID~\cite{theis2015note}, which rely primarily on pixel-level comparisons, our approach leverages the structural properties of SVG, an XML-based vector graphics format, to capture both visual fidelity and underlying XML code consistency. SVG representations are not merely visual outputs; they are composed of structured commands, attribute sets, and rendering logic that encode information often missed by purely perceptual metrics.

Our metric incorporates both macroscopic visual alignment and microscopic structural correspondence. At the macroscopic level, we compute a global Intersection-over-Union (IoU) score to quantify the overlap of visual shapes. At the microscopic level, we perform a comprehensive analysis of SVG code. This begins with preprocessing steps such as center alignment and scale normalization, followed by path extraction and the construction of a multi-dimensional similarity matrix. The matrix integrates several components, including path-wise IoU, similarity to color attribute, and spatial relationship metrics. We employ the Hungarian algorithm to perform optimal path-level matching, ensuring robust structural alignment.Beyond direct path matching, we further introduce two structural consistency modules: path count consistency and path order scoring. The latter evaluates the semantic fidelity of rendering by assessing the alignment of drawing order, which in SVG inherently determines visual stacking and rendering priority. 

The final PSS score is computed through weighted summation, where the macroscopic visual alignment score receives a weight of 0.6 and the microscopic structural correspondence score receives a weight of 0.4, balancing the importance of overall visual quality and structural precision.

\section{Generation}
\subsection{Text-to-SVG Generation}
\label{app:tts}
\noindent\textbf{Construction Details.} In the Text-to-SVG Generation task, each evaluation sample is constructed as a paired instance of a natural language caption and its corresponding SVG. To ensure that the textual descriptions exhibit high semantic coverage and expressive fidelity, we employ a semi-automated pipeline as follows.

First, we render the original SVG into a high-resolution raster image using the Cairosvg library. The rendered image is then passed to the Claude-3.7-sonnet~\cite{anthropic2023claude2} model along with a minimal prompt, for example, "Describe this image in brief". The model generates an initial image caption that is then manually reviewed and refined. This human-in-the-loop step ensures that each caption accurately captures the core visual elements of the graphic, including structural layout, component composition, color attributes, and stylistic traits. To maintain quality, outputs exhibiting semantic ambiguity, structural inaccuracies, or excessive token length are systematically filtered.

This hybrid pipeline combines the scalability of automatic generation with the precision of manual validation, resulting in a high-quality dataset of paired captions and SVGs suitable for model evaluation.

\noindent\textbf{Evaluation Metrics.} To systematically evaluate the performance of models on the Text-to-SVG task, we propose a three-dimensional evaluation framework encompassing perceptual quality, visual reconstructability, and semantic alignment:

\noindent\textbf{Perceptual Quality.} This dimension assesses subjective aesthetic appeal and user-level acceptability of generated SVG. We adopt two metrics:

\begin{itemize}[leftmargin=1em,topsep=5pt] 
\item \textbf{Aesthetic Score~\cite{schuhmann2022aesthetic}}: Computed using a pre-trained image aesthetics model, this score evaluates the visual pleasantness of the rendered SVG image.

\item \textbf{Human Preference Score (HPS)~\cite{wu2023human}}: Derived from a series of A/B testing rounds, this metric captures human preferences by comparing model-generated SVGs against reference graphics. The final score reflects the frequency with which participants favor the generated output.
\end{itemize}
\noindent\textbf{Visual Reconstructability.} This dimension evaluates how well the generated SVG replicates the structural code characteristics of the reference SVG. We design a suite of custom metrics PSS(cf. Appendix~\ref{app:evaluation metrics} for more details) that compare both SVG code and rendered output across subdimensions including path topology, geometric layout, and attribute encoding.

\noindent\textbf{Semantic Alignment.} This dimension measures the degree to which the generated SVG semantically aligns with the input caption. We employ two complementary metrics:

\begin{itemize}[leftmargin=1em,topsep=5pt] 
\item \textbf{CLIP Score}: The generated SVG is rendered and embedded alongside the input caption using a CLIP model. The cosine similarity between their embeddings quantifies semantic alignment.

\item \textbf{Relative CLIP Consistency (rCLIP)}: We propose this novel metric to evaluate the semantic degradation of the generated image relative to the ground truth. It is defined as:

\begin{equation}
\setlength{\abovedisplayskip}{6pt} %
\setlength{\belowdisplayskip}{6pt} %
\scalebox{0.95}{
$
rCLIP = 1 - \max \left( 0, \frac{CLIP(caption, GT) - CLIP(caption, GEN)}{CLIP(caption, GT)} \right)
$
}
\end{equation}

Here, $CLIP(caption, GT)$ denotes the CLIP similarity between the input caption and the ground-truth rendering, while $CLIP(ca-ption, GEN)$ measures similarity with the model-generated output. Higher rCLIP values indicate stronger semantic preservation.
\end{itemize}

\begin{tcolorbox}[
  colback=white,
  colframe=black,
  title=Prompt for Text-based Generation,
  fonttitle=\bfseries,
  coltitle=white,
  colbacktitle=gray!70!black,
  boxrule=0.5pt,
  arc=3mm,
  left=2mm,
  right=2mm,
  bottom=2mm,
  enhanced,
]
You are a professional SVG designer with extensive experience in vector graphics creation. Please generate the corresponding SVG code based on the user's description. \\
Provide your answer strictly in the following format: Answer: \{SVG code\}, providing the complete fixed code only in the {SVG code} position, without adding any explanations, comments, or other content.\\
Important Notes:\\ 
Your SVG must ONLY include <path> elements with "fill" and "d" attributes. Do not use any other SVG elements or attributes,and must use exactly this opening tag: <svg class="icon" viewBox="0 0 1024 1024" version="1.1" xmlns="http://www.w3.org/2000/svg">\\

Please generate an SVG code of: \{prompt\} 
\label{tab:prompt for text-based generation}
\end{tcolorbox}

\subsection{Image-to-SVG Generation}
\label{app:mts}
\noindent\textbf{Construction Details.}
In the Image-to-SVG Generation task, we aim to evaluate a model’s capability to generate structured SVG conditioned on both image and textual inputs. Each evaluation sample comprises three components: (1) a ground-truth SVG file, (2) its corresponding rendered raster image (generated via the CairoSVG library), and (3) a paired natural language description. The image-caption pairs are constructed following the same pipeline as in the Text-to-SVG task. 

By leveraging both visual and textual modalities, this task simulates realistic usage scenarios in which users express design intent through a combination of imagery and language. Consequently, it imposes a more challenging multimodal grounding requirement on generative models.

\noindent\textbf{Evaluation Metrics.} To comprehensively assess the performance of multimodal large language models (MLLMs) on this task, we propose two complementary categories of evaluation metrics, each targeting a distinct aspect of generation quality: perceptual similarity and visual reconstructability.

\noindent\textbf{Perceptual Similarity} These metrics quantify the visual fidelity of the generated SVGs relative to the reference graphics, both globally and locally:
\begin{itemize}[leftmargin=1em,topsep=5pt]
\item \textbf{Learned Perceptual Image Patch Similarity (LPIPS)~\cite{zhang2018unreasonable}}: Measures perceptual differences in deep feature space, capturing consistency in local texture and edge structures.
\item \textbf{Structural Similarity Index Measure (SSIM)~\cite{wang2004image}}: Evaluates similarity in terms of luminance, contrast, and structural alignment, with an emphasis on holistic visual reconstruction.

\item \textbf{DINO Score~\cite{oquab2023dinov2}}: Leverages semantic embeddings from the self-supervised DINO model to assess structural alignment between generated and reference images, particularly suited to evaluating contour preservation and compositional fidelity.
\end{itemize}

\noindent\textbf{Visual Reconstructability} Given the structural and syntactic nature of SVG graphics, we further introduce metrics that analyze both image-level and code-level alignment:

\begin{itemize}[leftmargin=1em,topsep=5pt] 
\item \textbf{MSE}: Computes pixel-wise differences between the rendered outputs of generated and reference SVGs within a spatially aligned viewport, incorporating boundary-aware weighting to emphasize contour fidelity.

\item \textbf{Path-Structure Similarity Score (PSS)}: Extracts and compares SVG path commands, attributes, and hierarchical organization to assess fine-grained geometric and syntactic alignment. (cf. Appendix~\ref{app:evaluation metrics} for more details)
\end{itemize}

Together, these perceptual and structural metrics provide a robust and reproducible evaluation framework for multimodal generative tasks, capturing both the visual plausibility and syntactic correctness of generated SVGs.

\begin{tcolorbox}[
  colback=white,
  colframe=black,
  title=Prompt for Image-based Generation,
  fonttitle=\bfseries,
  coltitle=white,
  colbacktitle=gray!70!black,
  boxrule=0.5pt,
  arc=3mm,
  left=2mm,
  right=2mm,
  bottom=2mm,
  enhanced,
]
You are a professional SVG designer with extensive experience in vector graphics creation. Please generate the corresponding SVG code based on the user's description and the provided image reference. \\
 Provide your answer strictly in the following format: Answer: \{SVG code\}, providing the complete fixed code only in the {SVG code} position, without adding any explanations, comments, or other content. \\
Important Notes: \\Your SVG must ONLY include <path> elements with "fill" and "d" attributes. Do not use any other SVG elements or attributes,and must use exactly this opening tag: svg class="icon" viewBox="0 0 1024 1024" version="1.1" xmlns="http://www.w3.org/2000/svg"> \\

Please generate an SVG code based on this description: \{prompt\} and the reference image I've provided. 
\label{tab:prompt for Image-to-SVG generation}
\end{tcolorbox}

\subsection{Style Transfer}
\label{app:st}
\textbf{Construction Details.} In the Style Transfer Generation task, we aim to evaluate the model's ability to perform cross-style transformations of SVG graphics while preserving both the underlying structure and semantic content. Given the context-length limitations of large multimodal models when processing long SVG sequences, we prioritize the selection of compact and semantically clear SVG samples during dataset construction. This ensures that the transformation process focuses on stylistic expression rather than structural reconstruction.

We define four representative target styles that span key axes of variation across mainstream visual aesthetics:

\begin{itemize}[leftmargin=1em,topsep=5pt] 
\item \textbf{3D Style}: Enhances depth and lighting effects, assessing the model’s ability to abstract and reconstruct complex rendering semantics.

\item \textbf{Line art}: Emphasizes outlines and linear representations, requiring the model to simplify visual elements while retaining structural integrity.

\item \textbf{Pixel art}: Mimics low-resolution rasterized appearances, testing the model’s capacity for detail compression under strict structural constraints.

\item \textbf{Cartoon Style}: Introduces dynamic and emotive visual semantics, such as cartoonish or anthropomorphic traits, highlighting the model’s ability in affective style transfer.
\end{itemize} 

To increase both the discriminability and difficulty of the Style transfer task, each sample is manually assigned a target style that is maximally divergent from the original SVG style. For instance, icon-like SVGs with well-defined geometric outlines are more often mapped to Pixel or Cartoon styles, while graphics involving shading or gradients are preferentially assigned to the Line art style. This strategy maximizes the stylistic shift in each task instance, thus enhancing the challenge for models and the effectiveness of evaluation.

Each task instance is composed of three elements: (1) an original SVG graphic, (2) a target style label (chosen from the four defined categories), and (3) a natural language instruction (e.g., please convert this graphic to pixel art style). This formulation not only provides a consistent prompt format, but also enforces the requirement that the output preserves semantic structure while completing the stylistic transformation, facilitating clearly aligned downstream evaluation.

\noindent\textbf{Evaluation Metrics.} We develop a two-tier automated assessment framework leveraging LLMs to quantify transfer quality from global and local perspectives.

\noindent\textbf{Global Ranking Evaluation.}
Based on the results of the multi-dimensional evaluation, as shown in Table~\ref{tab:overall ranking} we identify six high-performing models for subsequent holistic assessment: the proprietary models Claude-3.7-Sonnet~\cite{anthropic2023claude2}, Gemini-2.0-Flash~\cite{team2024gemini}, and GPT-4o~\cite{hurst2024gpt}, along with the open-source models DeepSeek-R1~\cite{deepseek2024r1}, QwQ-32B~\cite{yang2025qwen3}, and Qwen3-32B~\cite{yang2025qwen3}.

\begin{table}[h]
\centering
\small
\caption{Comparison of Overall Rank Evaluation }
\label{tab:overall ranking}
\begin{tabular}{c l}
\toprule
Models & Winrate(\%) \\
\midrule
Claude-3.7-Sonnet~\cite{anthropic2023claude2} & 61.54 \\
DeepSeek-R1(Reference Model)~\cite{deepseek2024r1} & 50.55 \\
Gemini-2.0-Flash~\cite{team2024gemini} & 49.28 \\
QwQ-32B~\cite{yang2025qwen3} & 47.06 \\
GPT-4o~\cite{hurst2024gpt} & 45.45 \\
Qwen3-32B~\cite{yang2025qwen3} & 40.26 \\
\midrule
\end{tabular}

\end{table}

The final evaluation focuses on three criteria: semantic preservation, stylistic fidelity, and visual quality. We employ the AlpacaEval framework using DeepSeek-R1~\cite{deepseek2024r1} as the reference model. The outputs of each competing model are compared against the reference model's results across all samples. As shown in Table~\ref{tab:overall ranking}, the final rankings are derived from aggregated win–loss statistics, sorted by win rate relative to the baseline.

\noindent\textbf{Local Automated Multi-dimensional Evaluation.} The stylized SVG outputs are first rendered into image format and subsequently scored on a scale of 1–5 by an evaluation model (with a score of 0 assigned to invalid or erroneous generations). We design five metric including Content Preservation (CP), Detail Fidelity (DF), Style Consistency (SC), Color Harmony (CH) and Composition Balance (CB). More details are shown in Table~\ref{tab:automatic score rubric}. For each dimension of evaluation, we define qualitative descriptors that correspond to each level of score. We adopt GPT-4o-mini~\cite{hurst2024gpt} as the evaluation model, which references the original image, the stylized image, and the descriptor texts to first generate detailed feedback and then assign a final score. To ensure consistency of evaluation, the model operates with a deterministic temperature setting of 0.05.

\begin{tcolorbox}[
  colback=white,
  colframe=black,
  title=Prompt for Automatic Evaluation of Style Transfer,
  fonttitle=\bfseries,
  coltitle=white,
  colbacktitle=gray!70!black,
  boxrule=0.5pt,
  arc=3mm,
  left=2mm,
  right=2mm,
  bottom=2mm,
  before skip=8pt,    %
  enhanced,
]
You are a fair judge assistant tasked with providing clear, objective feedback based on specific criteria, ensuring each assessment reflects the absolute standards set for performance. \\

Task Description:
An instruction (might include an Input inside it), a response to evaluate, and a score rubric representing a evaluation criteria are given. \\
1. Make a brief description of the style transfer that how it modifies the image 1 to image 2. \\
2. Write a detailed feedback that assess the quality of the response strictly based on the given score rubric, not evaluating in general. \\
3. After writing a feedback, write a score that is an integer between 1 and 5. You should refer to the score rubric. \\
4. The output format should look as follows: "Feedback: (write a feedback for criteria) [RESULT] (an integer number between 1 and 5)" \\
5. Please do not generate any other opening, closing, and explanations. \\

The instruction to evaluate:\\
transfer the provided Image 1 to \{style\} \\

Response to evaluate:\\
the given Image 2 \\

Score Rubrics:\\
\{rubric\}  \\

Feedback:
\label{tab:prompt for automatic evaluation of style transfer}
\end{tcolorbox}

\begin{tcolorbox}[
  colback=white,
  colframe=black,
  title=Prompt for Style Transfer,
  fonttitle=\bfseries,
  coltitle=white,
  colbacktitle=gray!70!black,
  boxrule=0.5pt,
  arc=3mm,
  left=2mm,
  right=2mm,
  bottom=2mm,
  enhanced,
]
You are a professional SVG designer with extensive experience in vector graphics creation. 
Your task is to perform a style transfer - recreate the provided reference SVG.
Maintain the basic structure and given semantic description of the SVG, but adjust it to the desired style.
Provide your answer strictly in the following format: Answer:\{SVG Code\}, without adding any explanations, comments, or other content. \\

Reference SVG to transform: \\
\{reference\_svg\}\\

Description of the reference SVG: \\
\{description\}\\

the style to transfer:\\
\{style\} 

\label{tab:prompt for style transfer}
\end{tcolorbox}

\begin{table*}[htbp]
\centering
\small
\caption{Details of Score Rubrics used in Automatic Evaluation }
\label{tab:automatic score rubric}
\begin{tabular}{c l}
\toprule
criteria description & Content Preservation: To what extent does the SVG conversion preserve the basic content and main elements of the original image? \\
score1 description & Essential elements are missing or severely distorted; core content is unrecognizable. \\
score2 description & Key elements are present but with noticeable omissions or alterations.\\
score3 description &Most main elements are preserved, though some secondary features may be simplified.\\
score4 description & All critical content is accurately retained with minor detail loss.\\
score5 description &Perfect preservation of all primary and secondary elements without compromise.\\
\midrule
criteria description & Detail Fidelity: To what extent does the SVG conversion preserve the details of the original image? \\
score1 description & Fine details are completely lost; output appears overly simplified or blurred. \\
score2 description & Basic details are visible but intricate features (e.g., textures, small text) are poorly rendered. \\
score3 description & Moderate detail retention; some high-frequency elements may lack precision. \\
score4 description & High-fidelity details with only subtle imperfections in complex areas. \\
score5 description & Pixel-perfect detail reproduction matching the original's complexity. \\
\midrule
criteria description & Style Consistency: Compared to the reference style, how well does the conversion match the target style? \\
score1 description & Style is inconsistent or contradictory to reference (e.g., line art vs. painterly). \\
score2 description & Partial style adherence with noticeable deviations in techniques/effects. \\
score3 description & Broadly matches reference style but lacks nuanced execution. \\
score4 description & Close stylistic alignment with minor variations in rendering methods. \\
score5 description & Seamless style replication that could pass as original artwork. \\
\midrule
criteria description & Color Harmony: How harmonious is the color combination in the conversion result? \\
score1 description & Clashing or jarring colors; poor contrast/balance distracts from content. \\
score2 description & Some discordant color pairs but maintains basic readability. \\
score3 description & Generally pleasing palette though certain hues may feel slightly off. \\
score4 description & Well-balanced colors with intentional aesthetic cohesion. \\
score5 description & Masterful color theory application enhancing visual appeal. \\
\midrule
criteria description & Composition Balance: How balanced is the visual composition of the conversion result? \\
score1 description & Unbalanced weighting creates visual tension or emptiness. \\
score2 description & Some elements feel misaligned or disproportionately emphasized. \\
score3 description & Adequate balance though certain areas could benefit from adjustment. \\
score4 description & Strong compositional flow with deliberate focal points. \\
score5 description & Expert-level layout adhering to design principles (e.g., rule of thirds, negative space). \\
\bottomrule
\end{tabular}
\end{table*}

\section{Understanding}
\label{app:understanding}
\noindent\textbf{Construction Details.} To systematically evaluate the cognitive capabilities of large vision-language models (VLMs) on SVG-based graphics, we propose an understanding task comprising two subtasks: Perceptual Understanding and Semantic Understanding. Given the current limitations of multimodal models in directly parsing SVG code, we adopt an indirect approach. Specifically, raw SVGs are rendered into standard bitmap images using CairoSVG, which are then fed into the Qwen2.5-VL-72B-Instruct model for automatic generation of multi-choice question-answer (QA) samples.

Perceptual QA focuses on low-level visual features of the graphic, including shape recognition, color identification, and relative spatial positioning. Representative questions include:
"Which of the following basic shapes is present in the image?",
"What is the dominant color of the figure?", or
"In which direction is the ellipse located relative to the center?"

Semantic QA targets higher-level semantic interpretation, such as the meaning, functionality, or plausible real-world use of the graphic. Example questions include:
"Which of the following scenarios is the image most likely to represent?" or
"Which software interface might use this icon?"

To ensure the quality of the QA samples, all generated questions and answer options undergo manual validation. This includes filtering out ambiguous phrasing and reconstructing distractors to guarantee that each question has a single correct answer and well-separated alternatives.

\noindent\textbf{Evaluation Metric.} We adopt accuracy as the sole evaluation metric for the Understanding Task. For each multiple-choice question, the model is required to select exactly one correct answer. Accuracy is computed as the ratio of correctly answered questions to the total number of questions within each subtask. This metric offers a straightforward and interpretable means of assessing model performance on both perceptual and semantic dimensions of visual understanding.

\begin{tcolorbox}[
  colback=white,
  colframe=black,
  title=Prompt for Generating Questions in Perceptual and Semantic QA,
  fonttitle=\bfseries,
  coltitle=white,
  colbacktitle=gray!70!black,
  boxrule=0.5pt,
  arc=3mm,
  left=2mm,
  right=2mm,
  bottom=1mm,
  enhanced,
  label={box:prompt-qa},
  before skip=4pt,
  after skip=4pt,
]
Please analyze the provided icon image and its caption, then generate 2 multiple choice questions (one for each category below). Each question should have four options (A, B, C, D) with exactly one correct answer. \\
Generate these two types of questions: \\
1. Perceptual Question: Focus on the visual features of the icon such as shapes, number of elements, or spatial arrangement.  \\
2. Semantic Question: Explore the meaning, function, or use-case of the icon. Focus on what the icon represents or where it might typically appear.  \\
Format requirements: - Question: [question text] Options: A) [option A]; B) [option B]; C) [option C]; D) [option D] \\
Answer: [correct option letter] \\
Guidelines: \\
- All questions must be directly grounded in the icon image and its caption \\
- All alternative options should be plausible but clearly distinguishable from the correct answer \\
- Ensure questions have varying difficulty levels appropriate to their category \\
- The correct answer should be objectively verifiable based on the image and caption 
\label{tab:prompt for Generating Questions for PQA and SQA}
\end{tcolorbox}

\begin{tcolorbox}[
  colback=white,
  colframe=black,
  title=Prompt for Perceptual and Semantic QA,
  fonttitle=\bfseries,
  coltitle=white,
  colbacktitle=gray!70!black,
  boxrule=0.5pt,
  arc=3mm,
  left=2mm,
  right=2mm,
  bottom=2mm,
  enhanced,
]
You are a svg analysis expert. Follow these steps carefully to answer the given multiple choice question. \\ 
Task instruction:\\
1. Answer the given multiple choice question below according to the svg code.\\
2.Output the answer in the format 'Answer: X' in the last line, where X is one of A, B, C, or D.\\

SVG Code:\\
\{svg\_image\} \\
Question:\\
\{question\} \\
Options:\\
\{options\_str\} \\

Important Notes:\\
- You should answer exactly the given multiple-choice question, DO NOT propose a new question.\\
- Your output in the last line must be strictly in the format 'Answer: X', where X is one of A, B, C, or D.\\

Now, answer the given question:
\label{tab:prompt for PQA and SQA}
\end{tcolorbox}

\section{Editing}

\subsection{Bug Fixing}
\label{app:bf}
\noindent\textbf{Construction Details.} The bug fixing task is designed to systematically evaluate a model’s ability to identify and repair syntactic anomalies in SVG. This task simulates common error scenarios that arise in real-world design workflows, focusing on three major categories of typical faults:

\begin{itemize} [leftmargin=1em,topsep=5pt]
\item\textbf{Tag Errors}: Structural issues in the XML hierarchy due to incorrect or malformed element tags.

\item\textbf{Path Command Errors}: Logical inconsistencies in shape rendering caused by invalid or misused path commands.

\item\textbf{Attribute Errors}: Disruptions in visual appearance resulting from incorrect attribute names or values.
\end{itemize}

Each task instance consists of the following components:

\begin{itemize} [leftmargin=1em,topsep=5pt]
\item\textbf{Corrupted SVG}: Automatically generated using our custom tool SVGErrorGenerator, which injects faults with clearly defined types and diverse manifestations, ensuring that the rendered output is visibly degraded.

\item\textbf{Ground-Truth SVG}: The original, error-free SVG file, used to evaluate the correctness of the model's output.
\end{itemize}

The error generation process supports fine-grained control over fault injection parameters, including error type, frequency, and random seed, allowing for balanced sample diversity and task difficulty. Supported error types include tag misspellings or omissions, illegal path command substitutions, and incorrect attribute keys or values.

\noindent\textbf{Evaluation Metrics.} To comprehensively assess the performance of models in this task, we adopt the accuracy of bug fixing as the core evaluation metric:

\begin{itemize} [leftmargin=1em,topsep=5pt]
\item\textbf{Accuracy (ACC)}: Measures the proportion of SVG outputs that are fully restored to a structurally and semantically correct state. Evaluation is based on a strict equivalence check against the ground-truth SVG to determine successful repairs.
\end{itemize}
This metric characterizes a model's end-to-end capability in the detect-locate-repair loop. Importantly, we emphasize that successful repairs must not only restore structural validity, but also recover the intended design semantics and rendering fidelity.

\begin{tcolorbox}[
  colback=white,
  colframe=black,
  title=Prompt for Bug Fixing Task,
  fonttitle=\bfseries,
  coltitle=white,
  colbacktitle=gray!70!black,
  boxrule=0.5pt,
  arc=3mm,
  left=2mm,
  right=2mm,
  bottom=2mm,
  enhanced,
]    
You are a professional SVG repair engineer with expertise in SVG standards and common error types. \\
Your task is to analyze submitted SVG code, precisely locate errors, and fix problems using the principle of minimal modification. \\
Please only modify the parts causing errors while keeping the rest of the code unchanged. \\
After fixing, return the complete corrected code in the following strict format: Answer:\{SVG code\}, providing the complete fixed code only in the {SVG code} position, without adding any explanations, comments, or other content. \\

BUG SVG: \\
\{bug\_svg\} 
\label{tab:prompt for bug fixing task}
\end{tcolorbox}
\subsection{Code Optimization}
\label{app:co}
\noindent\textbf{Construction Details.} The code optimization task is designed to evaluate the model’s ability to perform structural compression and syntactic refactoring of SVG source code while preserving its rendered appearance. Inspired by the optimization strategies employed by the widely adopted open-source tool SVGO~\cite{svgo}, we construct an automated dataset tailored for structure-aware SVG code optimization.

Each sample in the dataset comprises the following three components:

\begin{itemize} [leftmargin=1em,topsep=5pt]
\item \textbf{Optimization Instruction}: Expressed in natural language, explicitly prompting referencing the SVGO~\cite{svgo} principle the model to compress, simplify, and remove redundancies in the SVG code according to professional standards.

\item \textbf{Original SVG}: Our dataset, featuring diverse structures and frequently containing optimizable redundancies.

\item \textbf{Groud-Truth SVG}: Generated automatically via a custom Python interface to the SVGO~\cite{svgo} library, ensuring structural validity and visual fidelity with respect to the original SVG.
\end{itemize}

The dataset spans a wide range of graphical complexities, providing the model with realistic optimization scenarios that reflect practical code refinement tasks.

\noindent\textbf{Evaluation Metrics.} To comprehensively assess the performance of the model in two key dimensions: compression effectiveness and visual fidelity. We propose the following dual-metric evaluation scheme:

\begin{itemize} [leftmargin=1em,topsep=5pt]
\item\textbf{Compression Code Ratio (CCR)}: This metric quantifies the reduction in byte size of the optimized SVG relative to the original SVG. It serves as a proxy for the model’s ability to compactly restructure the code without altering its visual rendering. The compression ratio is defined as:
\setlength{\abovedisplayskip}{6pt}
\setlength{\belowdisplayskip}{6pt}
\begin{equation}
CCR = (1-\frac{ \text{Optimized Size}}{\text{Original Size}}) \times 100\%
\end{equation}

\item\textbf{MSE}: To ensure visual fidelity, we compute the pixel-wise Mean Squared Error (MSE) between rasterized images of the original SVG and the model-optimized SVG. This metric captures rendering discrepancies and evaluates whether the visual output remains perceptually consistent after optimization.

\end{itemize}

To enhance sensitivity, we additionally report the MSE trend under varying compression levels, allowing for finer-grained analysis of whether aggressive compression leads to visual degradation. This enables detection of trade-offs where structural compactness may come at the cost of fidelity.

The core objective of this task is to encourage models to achieve structural compression without compromising visual consistency. Hence, optimized outputs with higher compression ratios and lower MSE scores are considered indicative of stronger optimization capabilities.

\begin{tcolorbox}[
  colback=white,
  colframe=black,
  title=Prompt for Code Optimization Task,
  fonttitle=\bfseries,
  coltitle=white,
  colbacktitle=gray!70!black,
  boxrule=0.5pt,
  arc=3mm,
  left=2mm,
  right=2mm,
  bottom=2mm,
  before skip=8pt,    %
  enhanced,
]    
You are an advanced SVG optimization expert, skilled at maximizing compression and optimization of SVG code while maintaining visual consistency and ensuring code correctness.\\
Please optimize the user-provided SVG code according to the following core principles:\\

1. Remove metadata and editor information\\
- Clear metadata, comments, and unnecessary attributes generated by design software\\
- Remove hidden elements and empty tags\\
2. Path optimization\\
- Simplify path data, reduce control points\\
- Lower decimal precision (1-2 places is usually sufficient)\\
- Merge similar paths\\
3. Attribute and style processing\\
- Remove redundant or default attribute values\\
- Merge duplicate styles\\
- Optimize color representation (e.g., \#000 instead of \#000000)\\
4. Structure optimization\\
- Remove unnecessary grouping and nesting\\
- Optimize IDs and class names\\
- Ensure viewBox is set correctly\\
5. Compression and fine-tuning\\
- Remove unnecessary whitespace and units\\
- Use short commands instead of long format commands\\

After optimization, please strictly return the complete optimized code in the following format:Answer: \{SVG code\},Provide the complete optimized code only in the {SVG code} position, without adding any explanations, comments, or other content.\\

Here is the original SVG to optimize:\\
\{origin\_svg\}
\label{tab:prompt for Code Optimization task}
\end{tcolorbox}
\subsection{Style Edting}
\label{app:se}
\noindent\textbf{Construction Details.} The SVG Editing Task is designed to evaluate the capability of large language models (LLMs) to perform localized and controllable modifications on structured graphics. We construct a synthetic dataset containing diverse types of editing operations. Each sample consists of three components: a natural language editing instruction, the original SVG graphic, and the corresponding ground-truth SVG after modification.

The editing instructions span a variety of common transformation types, including but not limited to:
\begin{itemize}[leftmargin=1em,topsep=5pt] 
\item Element-level color modifications (e.g., fill color, stroke color)

\item Geometric transformations (e.g., rotation, translation, scaling)

\item Stylistic enhancements and adjustments (e.g., blur filters, gradients, stroke width)
\end{itemize}

All samples are automatically generated via programmatic scripts to ensure semantic clarity, structural validity, and compatibility with automated evaluation pipelines. To support fine-grained manipulation of SVG elements, we develop a custom SVG editing toolkit based on Python and lxml, enabling precise control over element-level modifications. Each task instance pairs a well-defined natural language instruction with a reproducible target SVG, ensuring consistency and objectivity across the evaluation dataset.

\noindent\textbf{Evaluation Metrics.} Traditional image-level metrics (e.g., Mean Squared Error, MSE) are often insufficiently sensitive for SVG editing tasks, as they may not capture fine-grained structural changes and can be misleading due to global rendering differences. To address this, we introduce three metric evaluation schemes that emphasize both structural fidelity and perceptual relevance:

\begin{itemize}[leftmargin=1em,topsep=5pt] 
\item \textbf{Relative Levenshtein Distance (RLD)}: Measures the minimal structural modification cost between the model-generated SVG and the ground-truth SVG, computed over the raw SVG markup. This captures the syntactic efficiency of the model's edits.

\item \textbf{Relative Mean Squared Error (rMSE)}: A structure-aware ratio metric that quantifies the degree of visual correction (or degradation) relative to the original input. It is defined as:
\setlength{\abovedisplayskip}{6pt}
\setlength{\belowdisplayskip}{6pt}
\begin{equation}
rMSE = \sqrt{1 - \min\left( 1, \frac{MSE(gen, gt)}{MSE(gt, ori)} \right)}
\end{equation}

where $gen$ is the model's rendered output, $gt$ is the rendered ground-truth SVG, and $ori$ is the rendered original SVG. A larger $rMSE$ indicates that the model output more closely aligns with the intended target and represents a significant structural improvement over the original.

\item \textbf{Accuracy}: We employ accuracy as the primary evaluation metric to assess models' proficiency in executing specific editing operations correctly, where accuracy is computed as the proportion of successful task completions over the total number of editing attempts across all test samples.

\end{itemize}

This evaluation framework jointly captures the syntactic and semantic accuracy of model edits, offering fine-grained insight into model performance on real-world SVG editing tasks and enhancing the reliability of comparative analysis.

\begin{tcolorbox}[
  colback=white,
  colframe=black,
  title=Prompt for Style Editing Task,
  fonttitle=\bfseries,
  coltitle=white,
  colbacktitle=gray!70!black,
  boxrule=0.5pt,
  arc=3mm,
  left=2mm,
  right=2mm,
  bottom=2mm,
  enhanced,
]    
You are a professional SVG editing engineer with extensive experience in SVG editing. \\
Your task is to receive SVG code and modification requests from users, and make precise modifications according to their instructions. Please only modify the parts specified by the user, keeping the rest of the code unchanged. After fixing, return the complete corrected code in the following strict format: Answer:\{SVG code\}, providing the complete fixed code only in the \{SVG code\} position, without adding any explanations, comments, or other content. \\

Here is the original SVG:\\
\{original\_svg\} \\

Edit command:\\
\{edit\_command\} 
\label{tab:prompt for style editing task}
\end{tcolorbox}

\end{document}